\documentclass{siamart251216}
\usepackage[T1]{fontenc}
\usepackage{amsfonts,amssymb,booktabs,array,microtype,graphicx}
\usepackage{mathtools}
\usepackage{enumitem}
\usepackage{algpseudocode}
\usepackage{xurl}
\usepackage{needspace}
\usepackage{float}
\usepackage{tikz}
\usetikzlibrary{arrows.meta}
\graphicspath{{figures/}}
\hypersetup{pdftitle={Tensor network structured neural networks},
  pdfauthor={Rizheng Huang and Joe Kileel},colorlinks=true,linkcolor=black,citecolor=black,urlcolor=black,
  filecolor=black}
\newcommand{\contract}{\mathbin{\bullet}}
\newcommand{\jointcontract}{\mathop{\bullet}\nolimits}
\newcommand{\cT}{\mathcal T}
\newcommand{\tn}{\mathbf T}

\newcommand{\Can}{\mathcal C}
\newcommand{\R}{\mathbb R}
\newcommand{\E}{\mathbb E}
\newcommand{\norm}[1]{\left\lVert#1\right\rVert}
\newcommand{\inner}[2]{\left\langle#1,#2\right\rangle}

\newcommand{\softmax}{\operatorname{softmax}}

\newsiamremark{remark}{Remark}
\setlist[itemize]{itemsep=2pt,topsep=3pt,parsep=0pt}

\makeatletter
\renewcommand{\paragraph}{\@startsection{paragraph}{4}{\z@}%
  {.75\baselineskip \@plus .15\baselineskip \@minus .1\baselineskip}%
  {.2\baselineskip}%
  {\normalfont\normalsize\bfseries}}
\makeatother
\title{Tensor network structured neural networks}
\author{Rizheng Huang\thanks{Oden Institute for Computational Engineering and Sciences,
University of Texas at Austin, 
(\email{rh42322@my.utexas.edu})}
\and Joe Kileel\footnotemark[1] \thanks{Department of Mathematics,
University of Texas at Austin, (\email{jkileel@math.utexas.edu})}}
\headers{Tensor network structured neural networks}{R. Huang and J. Kileel}

\begin{document}
\maketitle
\begin{abstract}
We construct neural networks that learn maps between tensor networks on
general graphs, and demonstrate scalable, accurate, and efficient
prediction in the {tensor train (TT)} setting.
Each layer updates the tensor cores through learned contractions and
nonlinear maps, and the resulting cores form the next hidden state.
Tensor network interaction
networks (TNIN) combine local core interactions with tensor core attention
(t-attention) to connect distant cores.
In high-dimensional operator learning tests, an instantiation called the tensor train interaction
network (TTIN) achieves mean relative test errors of $0.077\%$ for
36-dimensional Hamilton-Jacobi-Bellman (HJB) evolution,
and $0.1555\%$ for 100-spin Glauber probability evolution. For the 256-spin
Ising chain, density matrix renormalization group (DMRG) initialized by
TTIN and followed by even-parity projection reaches $99.50\%$ mean test fidelity.
At comparable accuracy, TTIN-based pipelines have significantly lower end-to-end
latency than numerical solvers for HJB, Ising, and Glauber. For example, for 
256-spin Ising at batch size 32, the TTIN-initialized pipeline achieves
a $185\times$ speedup over a product state counterpart.
We also present a tensor train CNN (TT-CNN) as another instantiation of
our framework. On a benchmark for cross-subject activity recognition from
multichannel time series data, TT-CNN achieves state-of-the-art
mean test accuracy under source-domain validation, surpassing a
comparably sized traditional CNN by 7.53 percentage points.

\end{abstract}

\section{Introduction}\label{sec:introduction}
Many scientific and engineering applications require repeated evaluation of maps from
input data to solutions. 
Examples include solution operators
for families of {PDEs}
\cite{kovachki2023neuraloperator,lu2021deeponet} and maps from
Hamiltonians to their ground states \cite{schollwock2011}.
Learning nonlinear maps from data can reduce the cost of repeated numerical
solves.  In challenging cases, both the input and solution data are very high-dimensional.

On the other hand, tensor networks (TNs) 
provide a common framework for efficiently representing high-dimensional  data~\cite{liu2018multimodal,cao2024mritensor,schollwock2011}.
They can enable the curse of dimensionality for high dimensional data to be beaten in situations where data is structured or different parts of data interact with only certain other parts \cite{hackbusch2012tensor,TensorTextbook}.

To learn maps between high-dimensional spaces, in this paper we construct \emph{neural networks based on
nonlinear interactions among tensor cores}. 
Each hidden state is a
tensor network, and the learned updates act directly on the cores without reconstructing the corresponding dense tensors.
The construction applies to tensor networks on general graphs and
supports different interaction mechanisms.
More specifically, we propose 
encoder-backbone-decoder architectures of the form: %
\begin{equation}\label{eq:intro_architecture}
 \widehat{\mathcal F}_\theta:\quad
 x\xrightarrow{\mathcal E}
 \underbrace{\tn^{(\mathrm{in})}\to\cdots\to
   \tn^{(\ell)}\xrightarrow{{\Psi^{(\ell)}}}\tn^{(\ell+1)}
   \to\cdots\to\tn^{(\mathrm{out})}}_{\text{tensor network backbone }\mathcal N_\theta}
 \xrightarrow{\mathcal D}\widehat y.
\end{equation}
Each layer update map ${\Psi^{(\ell)}}$ couples the cores of the tensor network
$\tn^{(\ell)}$ through learned contractions and updates them nonlinearly.
The encoder $\mathcal E$ builds the input tensor network, and the
decoder $\mathcal D$ converts the final cores into the output required
by the task.
Our architectures scale to large problems and support accurate, efficient
prediction across a range of challenging tasks.

\subsection{Motivation}
Scientific computing often involves high-dimensional objects, e.g., in
operator learning~\cite{khoo2025correlations} and uncertainty
quantification~\cite{dolgov2019parametric}. Dense tensor
representations of these objects suffer from the curse of
dimensionality. For example, discretizing a function of $d$ variables
with $m$ grid points per variable produces a tensor with $\mathcal{O}(m^d)$ entries.
Tensor networks can reduce this cost by representing the object as a
contraction of smaller cores. A tensor network with $N$ cores stores $\mathcal{O}(Nm^p)$ entries, where $p$ and $m$ upper-bound the core order and mode
size.  So keeping the layer states inside a neural network in compact
tensor network form offers a way to reduce storage and for each layer to compute its
update directly from the smaller cores.

Besides compression considerations, it is also often the case that data has a natural tensor structure.  This arises through the discretization of multivariate functions or probability densities, or in such domains as sensor data
analysis~\cite{barshan2010dsads} and recommendation
systems~\cite{karatzoglou2010multiverse}. Tensor modes make
the organization of the input explicit. By contrast, vectorization combines
modes into a single index, making their roles in the input implicit.  %
Updating tensor network states through core interactions lets the
network use the input structure directly, potentially making the learning problem more tractable.   %

Together, these points motivate the general form of neural network architecture in
\cref{eq:intro_architecture}. At each layer of the backbone in \cref{eq:intro_architecture}, the cores store the state
compactly, and their connections guide the interactions that produce
the next layer's  tensor network.

\subsection{Related work}
Several prior works have used tensor networks to reduce the number of parameters in a learned map.
In particular, Novikov et al.\ parameterize fully connected layer weights
in {TT} form and train the cores~\cite{novikov2015tensorizing}. Stoudenmire
and Schwab use a trainable tensor network that contracts with a product
of local input features to produce class scores, combining tensor networks with kernel methods rather than neural networks~\cite{stoudenmire2016supervised}.

Prior work has also explored how tensor cores can carry features or predictions that vary with
the input. Maruhashi et al.\ use a learned target core to guide the
Tucker decomposition of each sample, giving the classifier a compact
representation of multiway relations~\cite{maruhashi2018deeptensor}.
While we were writing up our results, Grimm and Lindsey released work that uses a transformer to process tokens encoding
Hamiltonian parameters and map the final tokens to the cores of
a ground-state wavefunction. Their output representation makes it
possible to estimate observables via tensor network
contractions~\cite{grimm2026foundation}.

In addition,  researchers have exploited tensor structure to determine the operations within a layer.
Kossaifi et al.\ reduce the dimensions of an activation tensor
through learned modewise maps~\cite{kossaifi2020tensorregression}. Sharir et al.\ construct
feedforward networks that approximate the evaluation of a
tensor network quantum state by following a  contraction
scheme~\cite{sharir2022neuralcontractions}. A closer precedent for
updating a tuple of cores is the TT variant of Chen et al., which
augments a low-rank TT chain with a core for a sparse refinement so
that the two components can interact~\cite{chen2026dctnn}.
Simultaneously with the present work, Wang et al. introduce the tensor neural operator (TEN-NO) architecture, which learns  solution maps for time evolution and parametric equations in {TT} form, using TT-SVD truncation to keep the ranks of the updated states within prescribed bounds~\cite{wang2026}.

Fully exploiting tensor network structure during architecture design, training and inference, as well as enabling hidden states represented by tensor networks on general graphs,  all remain. Training and inference need not require SVD-based truncation, which may be expensive. %
Further, each neural network layer can specify which
input cores  contribute to each output core, and how to combine
them to produce the next tensor network state. 
We address these
outstanding issues in the present work, and demonstrate  competitive numerical results in a variety of applications. %

\subsection{Our contributions}
Our main contribution is that we give a unified construction for tensor network structured neural
networks, and instantiate some highly performant architectures for learning maps between high-dimensional spaces. In each layer the state is represented as a tensor network on a
 graph, and we form the next tensor network through direct core
interactions and nonlinear updates. 
Tensor network structured layers form the backbone
in \cref{eq:intro_architecture}, so that the encoder and decoder components can be
tailored to the inputs and outputs of each domain or task. Using this framework, we introduce several particular neural network architectures. 
{TNIN combines} local
core interactions with global interactions through {t-attention}, which extends standard attention to a
tensor core version. 
When the layerwise graph is a path, TNIN gives {TTIN}. 
The same construction supports tensorized versions of
various standard popular architectures, like  transformers~\cite{vaswani2017attention} and CNNs.
To reduce the number of learned coefficients, we
express selected weight modes in smaller fixed bases, allowing us to
contract in coefficient space and reduce computation and intermediate
storage (\cref{subsec:backbone_construction}).  As in prior work on operator learning with traditional neural
networks~\cite{kovachki2023neuraloperator}, this design choice makes the
number of learnable parameters in these weight tensors independent of
the original dimensions of the selected modes when the basis sizes are
fixed.

For when problem structure is insufficient to fix a common input gauge,
we contribute a TT canonicalization procedure to resolve the ambiguity caused
by gauge symmetry. We prove that two real TTs
representing the same tensor yield identical canonicalized cores 
(\cref{thm:tt_canonical}). We extend the construction to trees in
\cref{sec:treecanon}. In experiments, training on
canonicalized inputs yields lower test errors than training on inputs
obtained where encoding was done by TT-SVD~\cite{oseledets2011tt} alone.
Canonicalizing inputs at inference also makes a fixed model's predictions
independent of the input gauge (\cref{subsec:canonicalization_study}).

Lastly, we contribute concrete implementations and  computational  experiments to test out our new architectures. We  compare TTIN with conventional networks on
{HJB} evolution to assess the effect of
direct core interactions on accuracy. The conventional networks use
vectorized versions of the same input cores. On the benchmark  ``Daily and Sports Activities Dataset'' (DSADS)~\cite{barshan2010dsads},
we compare TT-CNN with a standard 1D CNN for classification from sensor data. The tensor core-based models are more
accurate in both comparisons at similar parameter counts. TT-CNN achieves
{state-of-the-art (SOTA)} accuracy on HAROOD's DSADS cross-subject
splits~\cite{lu2026harood}, surpassing the strongest baseline reported
under source-domain validation.
We evaluate TTIN at increasing problem sizes to assess its scalability.
TTIN maintains low test errors up to 36 variables
in HJB and 100 spins in Glauber, and TTIN-initialized refinement
yields high fidelity up to 256 spins in Ising. For HJB evolution, TTIN combines high
accuracy with fast inference, making it highly competitive with established
methods such as {the Fourier neural operator (FNO)}~\cite{li2021fno}, {DeepONet with proper orthogonal decomposition (POD-DeepONet)}~\cite{lu2022poddeeponet},
and {the tensor-train reduced-order model (TT-ROM)}~\cite{danis2025ttopinf}.
We also benchmark TTIN-based pipelines against established numerical solvers
at comparable accuracy to quantify latency savings and estimate how
many inference batches are needed to recover training time.
{To compare methods that use different internal representations,
we measure end-to-end latency from the same inputs to final outputs in a
common format (at inference time), including any conversions required by each method.}
The TTIN-based pipelines achieve substantially lower {latency}
than TT Cole--Hopf in HJB, DMRG~\cite{white1992dmrg} in Ising,
and {time-evolving block decimation (TEBD)}~\cite{vidal2004tebd} in Glauber. For HJB and Glauber, these
savings give TTIN a cost advantage in applications with moderate to
high demand for repeated solves. The Ising pipeline is especially
cost effective, requiring far fewer inference batches to recover
its training time (\cref{sec:experiments}).

\subsection{Notation}\label{subsec:notation}
{Unless stated otherwise, tensors are real. A tensor's order is its number
of modes (indices). For a representation graph $G=(V,E)$, let $N=|V|$
be the number of cores. The network $\tn=(\cT_i)_{i\in V}$ has fixed
index connections, and contracting its bonds gives the represented tensor
$\cT$. The indices $(\ell)$ and $(\ell+1)$ denote consecutive layers.
Tildes mark fixed-basis coefficients, and hats mark predictions.
For an order-$p$ tensor $\mathcal A$, the product $\mathcal A\times_k U$,
$U\in\R^{q\times d_k}$, changes mode $k$ from dimension $d_k$ to $q$.
We write $\mathcal A\mathop{\times}_{k\in J}U_k$ for the products along
modes in $J$ and $\mathcal A\times_{1:p}(U_1,\ldots,U_p)
:=\mathcal A\times_1 U_1\cdots\times_p U_p$ for products along all modes.
The contraction $\mathcal A\contract_{I,J}\mathcal B$ pairs equal-dimensional
modes in the ordered lists $I,J$. The remaining modes retain their order,
with those of $\mathcal A$ first. For several tensors,
$\jointcontract_{\Pi}(\mathcal A_1,\ldots,\mathcal A_s)$ contracts the pairs
in $\Pi$, leaving the other modes free. We omit subscripts when the pairs
are clear.
Full contraction of tensors of the same shape gives the Frobenius inner
product. We use $\inner{\cdot}{\cdot}$ for Euclidean and Frobenius inner
products and $\norm{\cdot}_2$, $\norm{\cdot}_F$ for their respective norms.
An unsubscripted norm is Euclidean for vectors and Frobenius for matrices
and tensors.}
\Needspace{5\baselineskip}
\section{A General Construction With Tensor Networks}
\label{sec:framework}
\subsection{Preliminaries}\label{subsec:preliminaries}
A tensor network  uses a finite undirected \textit{representation graph}
$G=(V,E)$. Each vertex carries a tensor core and may have open legs.
At vertex $i\in V$, the core
$\cT_i\in\R^{d_{i,1}\times\cdots\times d_{i,p_i}}$
has $p_i$ modes, with dimension $d_{i,k}$ along mode $k$.
Each edge pairs two modes of equal dimension, one from each incident
core. Their shared index is a \emph{bond index}.
An open leg corresponds to an unpaired mode, an \emph{open index}
of the represented tensor. We fix which edge or open leg corresponds
to each core axis, as well as the order of the open indices in the
represented tensor. Contracting the bond indices associated with
all edges in $E$ gives
\begin{equation}\label{eq:reconstruction}
 \cT=\jointcontract_E(\tn),
\end{equation}
where $\tn=(\cT_i)_{i\in V}$.  The cost of this contraction depends on the graph, index dimensions,
and contraction order.  In the neural networks developed below, there is one TN per layer, whose representation graph is denoted  $G^{(\ell)}=(V^{(\ell)}, E^{(\ell)})$ for layer $\ell$.

We remark that tensor networks include familiar tensor decompositions. 
A Tucker decomposition \cite{jin2024scalable} has
a central tensor core connected to mode matrices.  A tree network \cite{shi2006classical} uses
a connected acyclic graph. The canonical polyadic decomposition (CP) \cite{zhang2023moment}
can be written with mode matrices and a fixed superdiagonal core. 
A TT uses a path graph \cite{oseledets2011tt}, where an order-$N$ tensor is represented by a chain
\begin{equation}\label{eq:tt}
 \cT(j_1,\ldots,j_N)=T_1(j_1)\cdots T_N(j_N),\qquad
 \cT_i\in\R^{r_{i-1}\times m_i\times r_i},\quad r_0=r_N=1.
\end{equation}
Here $T_i(j_i)$ is a matrix slice of $\cT_i$ at open index
$j_i\in\{1,\ldots,m_i\}$. The bond dimensions $r_i$ bound the TT ranks.

\subsection{Encoder-backbone-decoder architecture}\label{subsec:encoder_decoder}
To reuse the backbone design across tasks, we compose an encoder
$\mathcal E$, a tensor network backbone $\mathcal N_\theta$, and a decoder
$\mathcal D$. For a target map
$\mathcal F:\mathbb X\to\mathbb Y$ between high-dimensional input and output spaces $\mathbb X$ and $\mathbb Y$, the approximation is
\begin{equation}\label{eq:architecture}
 \widehat{\mathcal F}_\theta(x)
   =\mathcal D\!\left(\mathcal N_\theta(\mathcal E(x))\right).
\end{equation}
The encoder and decoder can contain fixed algorithms and/or learned maps;
we suppress their trainable parameters in the notation.  Parameters $\theta$ are for the backbone.

\begin{enumerate}[label=\arabic*.,leftmargin=*,itemsep=2pt,topsep=2pt,parsep=0pt,partopsep=0pt]
\item \textbf{Encoder $\mathcal E$.} The encoder constructs the input tensor
network from raw-data arrays, e.g., by TT-SVD or tree tensor network singular value decomposition (TTN-SVD) \cite{hackbusch2012tensor}, or uses an analytic or
supplied tensor network. The encoder can also preprocess the input
to facilitate learning, for example through canonicalization
(\cref{subsec:canonicalization}), feature extraction, or mapping the
cores to the dimensions required by the backbone.
\item \textbf{Backbone $\mathcal N_\theta$.} Once the input network is encoded,
the backbone transforms it through nonlinear interactions among tensor
cores (\cref{subsec:backbone_construction}). \Cref{sec:instances}
develops specific interaction architectures.  Certain cores in one layer interact to produce each core in the subsequent layer of the backbone.
\item \textbf{Decoder $\mathcal D$.} The decoder converts the final tensor
network into an output in $\mathbb Y$, e.g., a factored tensor for regression
or class scores for classification. It may include postprocessing
to improve accuracy or satisfy task-specific requirements.
\end{enumerate}

\begingroup
\setlength{\intextsep}{4pt}
\setlength{\abovecaptionskip}{1pt}
\setlength{\belowcaptionskip}{0pt}
\begin{figure}[H]
\centering
\includegraphics[width=\linewidth]{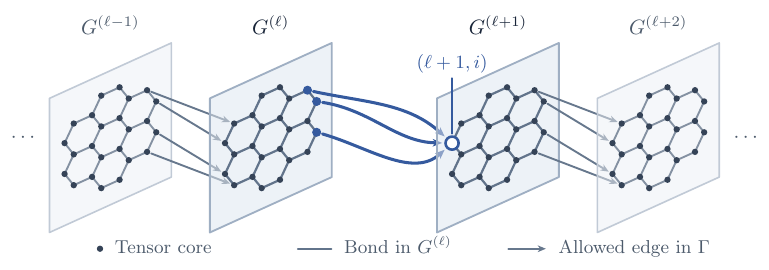}
\caption{{Example of layered computation graph $\Gamma$. The highlighted
arrows illustrate allowed dependencies of core $(\ell+1,i)$ on cores in
layer $\ell$.}}
\label{fig:layered_computation_graph}
\end{figure}
\endgroup
\enlargethispage*{0pt}
\vspace{-4pt}

\subsection{General construction of the backbone}\label{subsec:backbone_construction}
The backbone has a \textit{layered computation graph} $\Gamma$ {(\cref{fig:layered_computation_graph})}.  Its vertices are the various cores appearing in the backbone.
{Recall that each layer $\ell$ has its representation graph
$G^{(\ell)}=(V^{(\ell)},E^{(\ell)})$, with a core $\cT_i^{(\ell)}$ at each
vertex $i\in V^{(\ell)}$ and a bond for each edge in $E^{(\ell)}$. Vertex $i$
of $G^{(\ell)}$
corresponds to vertex $(\ell,i)$ of $\Gamma$. Meanwhile, edges in $\Gamma$ do not correspond to edges of  $E^{(\ell)}$; rather, arrows
$(\ell,j)\to(\ell+1,i)$ in $\Gamma$ specify which input cores can
contribute to core $i$ in the next layer. 
The computation graph $\Gamma$ controls the possible interactions between cores in the backbone.}

{To fully specify the backbone,
we also need an update rule for each output core. Given the layer state
$\tn^{(\ell)}=(\cT_j^{(\ell)})_{j\in V^{(\ell)}}$, we compute each output
core $\cT_i^{(\ell+1)}$ in four steps. When describing a single layer update below, we retain layer superscripts only
on the states and their cores.}

\begin{subequations}\label{eq:generalsubnetworkupdate}
\begin{enumerate}[label=\arabic*.,leftmargin=*,itemsep=4pt,topsep=4pt,parsep=0pt]
\item \textbf{Core contraction.}
{The update at core $i$ uses $n_i$ terms. For the
$k$th term, the update rule may choose any subset
$V_{i,k}$ of the input vertices $j$ for which $\Gamma$
contains the arrow $(\ell,j)\to(\ell+1,i)$. Let $E_{i,k}$
consist of all edges of the input representation graph $G^{(\ell)}$ with both endpoints in
$V_{i,k}$. Contracting the corresponding bonds gives}
\begin{equation}\label{eq:corecontraction}
 {\mathcal X_{i,k}}=
    \jointcontract_{{E_{i,k}}}\!\left(
       (\cT_j^{(\ell)})_{j\in{V_{i,k}}}\right).
\end{equation}
{This contraction is multilinear in the selected cores, and it
returns the selected core when $\lvert V_{i,k}\rvert=1$.}

\item \textbf{Weighted contraction.}
{The mode pairings $\Pi_{i,k}$ specify how $\mathcal X_{i,k}$
contracts with weight tensors $\mathcal W_{i,k,1},\ldots,\mathcal W_{i,k,s}$ . We choose the weight dimensions so
that the result can be reshaped to the prescribed shape of the output
core $\cT_i^{(\ell+1)}$, giving}
\begin{equation}\label{eq:weightedcontraction}
 {\mathcal S_{i,k}}=
    {\operatorname{reshape}_i}\!\left[
       \jointcontract_{{\Pi_{i,k}}}\!\left({\mathcal X_{i,k}},
         {\mathcal W_{i,k,1},\ldots,\mathcal W_{i,k,s}}\right)
    \right].
\end{equation}
{For fixed weights, this map is linear in $\mathcal X_{i,k}$.
The pairings may also connect modes of different weight tensors, allowing
the weights themselves to form a tensor network.}

\item \textbf{Linear combination.}
{Scalar coefficients $\alpha_{i,k}$ weight the terms
$\mathcal S_{i,k}$ from Step~2. Summing the weighted terms gives the
increment}
\begin{equation}\label{eq:coreaggregation}
 {\Delta_i}=\sum_{k=1}^{{n_i}}{\alpha_{i,k}}\,
       {\mathcal S_{i,k}}.
\end{equation}
{The coefficient $\alpha_{i,k}$ can be fixed at $0$ to omit
the term $\mathcal S_{i,k}$ or at $1$ to include it in the sum. It can
also be a trainable scale factor or depend on the cores of $\tn^{(\ell)}$
permitted by $\Gamma$. For example, in the t-attention layer of
subsection~\ref*{subsec:ttinattention}, we apply a softmax to scores
between the query from core $i$ and the keys from the input cores.}

\item \textbf{Core update.}
{At each output vertex $i$, the update map $\Phi_i$ uses the
increment $\Delta_i$ from Step~3 and may also use the input cores permitted
by $\Gamma$. The resulting core is}
\begin{equation}\label{eq:coreupdate}
 \cT_i^{(\ell+1)}={\Phi_i}(\tn^{(\ell)},{\Delta_i}).
\end{equation}
{For example, if $\Gamma$ contains the arrow
$(\ell,i)\to(\ell+1,i)$ and the two cores have the same shape, we can use the
residual update
$\Phi_i(\tn^{(\ell)},\Delta_i)=\cT_i^{(\ell)}+\sigma(\Delta_i)$.
Here a nonlinear activation function $\sigma$ acts entrywise on $\Delta_i$. This
increment has the shape of one output core and can have far fewer entries
than the dense tensor represented by $\tn^{(\ell+1)}$, thus entrywise operations can be efficient.}
\end{enumerate}
\end{subequations}

Steps~1 and~2 present core-core contractions before
core-weight contractions. In practice, we use associativity to reorder
the contractions while preserving the mode pairings and output index
order. A contraction order that reduces the operation count and peak
memory use accelerates training and inference. Beyond the savings
from reordering, batching compatible contractions across core updates
offers a further speedup on GPUs because all updates read the same input
layer.

A dense weight tensor with $q$ modes of dimension $m$ also requires
storage for $m^q$ trainable entries.
The architecture typically fixes $q$, while $m$
can grow with the problem size. For those modes, we use fixed bases
with a constant number of vectors in the contraction weights and
in the tensors that parameterize the coefficient functions ${\alpha_{i,k}}$.
\begin{samepage}
For a weight tensor $\mathcal W$, let $J$ be the set of modes represented
in fixed bases.
For each $k\in J$, take a fixed basis matrix $B_k\in\R^{d_k\times K_k}$
with $K_k\leq d_k$ and write
\begin{equation}\label{eq:fixedbasisweights}
 \mathcal W=\widetilde{\mathcal W}
      \mathop{\times}_{k\in J}B_k.
\end{equation}
\end{samepage}
\!\!Only $\widetilde{\mathcal W}$ is learned.  The fixed basis matrices $B_k$ may correspond, e.g., to evaluations of Fourier modes (among other possibilities).
For each $k\in J$, its mode
dimension is $K_k$, and the corresponding vectors of $\mathcal W$ lie
in the column space of $B_k$. Using $K\ll m$ basis vectors in each of
the $q$ modes of the dense tensor leaves $K^q$ trainable
coefficients, independent of $m$.  In most applications this has the interpretation that the number of trainable parameters is independent of the discretization resolution.

To perform the weighted contraction in \cref{eq:weightedcontraction}
in coefficient space, let one input mode and one output mode
of $\mathcal W$ use the same fixed basis matrix $B\in\R^{m\times K}$ with
orthonormal columns ($B^\top B=I_K$).
The pairings $\Pi$ match the input mode with mode $a$ of $\mathcal X$.
The output mode stays free at position $b$ of the result. By associativity,
\begin{equation}\label{eq:fixedbasiscontraction}
 \jointcontract_{\Pi}(\mathcal X,\mathcal W)
 =
 \jointcontract_{\Pi}\!\left(
   \mathcal X\times_a B^\top,\,
   \widetilde{\mathcal W}\right)\times_b B.
\end{equation}
Contracting $\mathcal X\times_a B^\top$ with $\widetilde{\mathcal W}$
uses the reduced mode dimension $K$ and avoids forming the dense weight
$\mathcal W$. To reduce the computation and storage needed to form
${\mathcal X_{i,k}}$, we can project the modes that stay free
on their individual cores before the contraction in \cref{eq:corecontraction}. For terms
with common output bases, we reduce the cost of addition by taking
the weighted sum of their smaller coefficient tensors. By linearity,
one expansion of the sum gives their contribution to the increment
${\Delta_i}$ in \cref{eq:coreaggregation} before applying ${\Phi_i}$.
\Cref{subsec:basis_analysis} describes the implementation.
\section{Neural Architectures on Tensor Networks}\label{sec:instances}
Starting from \cref{eq:generalsubnetworkupdate}, we obtain different
architectures by varying the input core sets, coefficients, or patterns
of weight sharing. Each can use the fixed basis parameterization as in
\cref{eq:fixedbasisweights}.

\subsection{Tensor core local interaction layers}\label{subsec:localinteraction}
Let ${\mathcal N_\rho(i)}$ contain the vertices at shortest-path distance
at most $\rho$ from $i$ in {the input representation graph}. For each term ${\mathcal S_{i,k}}$
in \cref{eq:weightedcontraction}, choose
${V_{i,k}\subseteq\mathcal N_\rho(i)}$.
Each term ${\mathcal S_{i,k}}$ has the current core's shape.
With fixed ${\alpha_{i,k}}\in\{0,1\}$ and entrywise activation $\sigma$,
\cref{eq:coreupdate} gives
\begin{equation}\label{eq:local_update}
 \cT_i^{(\ell+1)}=\cT_i^{(\ell)}
 +\sigma\!\left(\sum_{k=1}^{{n_i}}
          {\alpha_{i,k}}\,{\mathcal S_{i,k}}\right).
\end{equation}

\subsection{Interaction networks with t-attention}\label{subsec:ttinattention}
After $L$ local layers with radius $\rho$
{on a fixed representation graph}, core $i$ can depend only on
inputs within graph distance $L\rho$. On a path with $\rho=1$, this
covers at most $2L+1$ input cores. To let distant cores interact in a
single layer, we introduce t-attention.

We use $h$ t-attention heads indexed by $m$. For core $i$, the learned tuples
${\mathbf W_{i,m}^{\mathrm{Q}}}$, ${\mathbf W_{i,m}^{\mathrm{K}}}$, and
${\mathbf W_{i,m}^{\mathrm{V}}}$ play the roles of ${W^{\mathrm{Q}}}$, ${W^{\mathrm{K}}}$, and ${W^{\mathrm{V}}}$ in
standard attention \cite{vaswani2017attention}. With one matrix per mode,
the projections of an order-$p$ core are
\begin{equation}\label{eq:attentionmaps}
 {\mathcal Q_{i,m}}=\cT_i^{(\ell)}\times_{1:p}\!{\mathbf W_{i,m}^{\mathrm{Q}}},
 {\mathcal K_{i,m}}=\cT_i^{(\ell)}\times_{1:p}\!{\mathbf W_{i,m}^{\mathrm{K}}},
 {\mathcal V_{i,m}}=\cT_i^{(\ell)}\times_{1:p}\!{\mathbf W_{i,m}^{\mathrm{V}}}.
\end{equation}
To take inner products between any query ${\mathcal Q_{i,m}}$
and key ${\mathcal K_{j,m}}$, we choose the matrix row dimensions to
give all queries and keys in head $m$ a common shape with $D_m$ entries.
We likewise give the values ${\mathcal V_{j,m}}$ a common shape so
they can be combined in a weighted sum. Cores of the same shape can
share projection tuples. If core orders differ, we can add modes of
dimension one to obtain a common order.

\begin{samepage}
The bias ${b_{ij,m}}$ encodes position or graph information.
A softmax over {the input vertices $j$} gives the t-attention coefficients ${\alpha_{ij,m}}$:
\begin{equation}\label{eq:attention}
 {\alpha_{ij,m}}
 =\softmax_j\!\left(
    \frac{\inner{{\mathcal Q_{i,m}}}{{\mathcal K_{j,m}}}_F}{\sqrt{D_m}}
    +{b_{ij,m}}\right),\quad
 {\mathcal Z_{i,m}}
 =\sum_{{j}}{\alpha_{ij,m}}{\mathcal V_{j,m}}.
\end{equation}
\end{samepage}
\!\!To form an increment with the shape of $\cT_i^{(\ell)}$, we use $s$
learned output tuples ${\mathbf W_{i,k,m}^{\mathrm{O}}}$ per head.
Each matrix in a tuple maps one mode of ${\mathcal Z_{i,m}}$ to
the corresponding mode of $\cT_i^{(\ell)}$. We sum the mapped outputs
to obtain
\begin{equation}\label{eq:attnoutput}
 {\Delta_i}
 =\sum_{m=1}^{h}\sum_{k=1}^{s}
   {\mathcal Z_{i,m}}\times_{1:p}{\mathbf W_{i,k,m}^{\mathrm{O}}}.
\end{equation}
After removing any added modes of dimension one, we use the increment
in \cref{eq:coreupdate}. For a residual connection, we add it to the
current core to obtain $\cT_i^{(\ell+1)}=\cT_i^{(\ell)}+{\Delta_i}$.
To identify the terms in \cref{eq:coreaggregation}, substitute
\cref{eq:attention} into \cref{eq:attnoutput}. By linearity of the mode
products, each input core $j$ and output tuple contributes
${\mathcal V_{j,m}}\times_{1:p}{\mathbf W_{i,k,m}^{\mathrm{O}}}$ with
coefficient ${\alpha_{ij,m}}$.

In block $b$, the composition $\mathcal L_b$ of local layers forms
features at each vertex. The t-attention layer $\mathcal A_b$ then lets
each core combine these features across vertices.
An $n$-block {TNIN} is
\begin{equation}\label{eq:tnin}
 \mathcal N_\theta
 = (\mathcal A_n\circ\mathcal L_n)
   \circ\cdots\circ(\mathcal A_1\circ\mathcal L_1).
\end{equation}

The same construction extends a range of conventional architectures to
tensor networks, with transformers and {CNNs} as two examples. In a tensor network transformer, t-attention
replaces standard attention, and tensor network operator-state
contractions replace the matrix-vector products in the feedforward
block. On a path, these contractions apply a matrix product operator
(MPO) to a matrix product state (MPS). In a tensor network CNN, each
kernel offset uses a tuple of matrices, one per mode of the corresponding
core. The convolution then sums the transformed cores over the window.
Our experiments focus primarily on TTIN, but also include a {TT-CNN} experiment in \cref{sec:dsads}.

\begin{samepage}
\subsection{\texorpdfstring{{TTIN}}{TTIN}}\label{subsec:ttin}
{TTs} are widely used in scientific computing
\cite{oseledets2011tt}, making them a natural starting point
for studying TNIN. Specializing \cref{eq:tnin} to a path graph gives a
{TTIN}. For simplicity, we use
one block ($n=1$) in our experiments, with t-attention following
the local layers.

For $3\leq i\leq N-2$, every interaction span has two exterior bonds,
and the local update is
\begin{equation}\label{eq:ttlocalupdate}
\begin{aligned}
 \cT_i^{(\ell+1)}&=\cT_i^{(\ell)}+\sigma\Bigl(
   \alpha_{i,1}\,\cT_i^{(\ell)}\times_1 L_{i,1}\times_2 W_{i,1}\times_3 R_{i,1}\\
 &\quad+\alpha_{i,2}^-\bigl[\mathcal W_{i,2}^-\contract
       (\cT_{i-1}^{(\ell)}\contract\cT_i^{(\ell)})\bigr]
       \times_1 L_{i,2}^-\times_3 R_{i,2}^-\\
 &\quad+\alpha_{i,2}^+\bigl[\mathcal W_{i,2}^+\contract
       (\cT_i^{(\ell)}\contract\cT_{i+1}^{(\ell)})\bigr]
       \times_1 L_{i,2}^+\times_3 R_{i,2}^+\\
 &\quad+\alpha_{i,3}\bigl[\mathcal W_{i,3}\contract
       (\cT_{i-1}^{(\ell)}\contract\cT_i^{(\ell)}\contract\cT_{i+1}^{(\ell)})\bigr]
       \times_1 L_{i,3}\times_3 R_{i,3}\Bigr).
\end{aligned}
\end{equation}
\end{samepage}
\!\!Before applying $L$ and $R$, we order the free modes of each bracketed
contraction as left bond, open mode, and right bond. The maps match the
bond dimensions to those of $\cT_i^{(\ell)}$. The supplement gives the
rules for terms whose spans reach a chain end.
\section{Computation and Experiment Settings}\label{sec:computation}
\subsection{Gauge symmetry and canonicalization}\label{subsec:canonicalization}
A tensor network has gauge symmetry because different choices of bond
coordinates can represent the same tensor. On an edge $e=(u,v)$ with
incident modes $k(u,e)$ and $k(v,e)$, an invertible matrix $M$ defines
the change of coordinates
\begin{equation}\label{eq:gauge}
 (\cT_u,\cT_v)\longmapsto
 \left(\cT_u\times_{k(u,e)}M,\,
       \cT_v\times_{k(v,e)}M^{-\top}\right).
\end{equation}

The factors cancel when the bond $e$ is contracted in
\cref{eq:reconstruction}. Since the backbone acts on cores, its
predictions can still depend on the gauge.

To give the backbone identical cores for different TT representations
of the same tensor, we need a common convention for bond coordinates.
Problem structure can supply this convention. For example, the analytic
MPO for the Ising Hamiltonian $H$ in
\cref{eq:ising} specifies its cores in fixed coordinates. When no common
TT coordinates are prescribed, the encoder can apply
\cref{alg:tt_canonicalization} to the supplied cores. We assess this
preprocessing in \cref{subsec:canonicalization_study}. The supplement
extends the construction from chains to trees
(\cref{sec:treecanon}).

In \cref{alg:tt_canonicalization}, we first sweep from left to right
with {QR factorizations} to make the left
prefixes isometric. We then sweep from right
to left, using the subroutine $\operatorname{FixSVD}$ to choose an orthonormal row space
basis $W_i$ for each endpoint unfolding $T_i$. Because the left prefix
is isometric, the remaining tensor determines the Gram matrix
$T_i^\top T_i$. The basis depends only on this matrix and the fixed
column order, so identical remaining tensors yield the same endpoint
basis. We give the specific rule in Appendix~\ref{sec:tt_canonical_proof}.

\begin{algorithm}[H]
\caption{TT canonicalization}
\label{alg:tt_canonicalization}
\begin{algorithmic}[1]
\Require $(\cT_1,\ldots,\cT_N)$ as in \cref{eq:tt}.
\For{$i=1,\ldots,N-1$}
  \State $(Q_i,R_i)\gets\operatorname{qr}\bigl(\operatorname{reshape}(\cT_i;r_{i-1}m_i,r_i)\bigr)$,
    \quad $r_i\gets\min(r_{i-1}m_i,r_i)$
  \State $\cT_i\gets\operatorname{reshape}(Q_i;r_{i-1},m_i,r_i)$,
    \quad $\cT_{i+1}\gets\cT_{i+1}\times_1 R_i$
\EndFor
\For{$i=N,\ldots,2$}
  \State $T_i\gets\operatorname{reshape}(\cT_i;r_{i-1},m_i r_i)$
  \State $W_i\gets\operatorname{FixSVD}(T_i)$,
    \quad $r_{i-1}\gets\operatorname{rank}(T_i)$
  \State $\cT_i^{\rm can}\gets\operatorname{reshape}(W_i^\top;r_{i-1},m_i,r_i)$,
    \quad $\cT_{i-1}\gets\cT_{i-1}\times_3(T_i W_i)^\top$
\EndFor
\State \Return $(\cT_1,\cT_2^{\rm can},\ldots,\cT_N^{\rm can})$
\end{algorithmic}
\end{algorithm}

Here $\operatorname{qr}$ is thin QR, and reshapes preserve the core mode order.
\Cref{alg:tt_canonicalization} defines $\Can$ for nonzero tensors.
For an input representing zero, set $\Can(\tn)=\mathbf 0_{\mathrm{TT}}$,
with bond dimensions equal to one, a zero first core, and $e_1$ at every other site.

\begin{theorem}[TT canonicalization]\label{thm:tt_canonical}
\upshape
If two {TTs} $\tn$ and $\tn'$ with
the same ordered open modes represent the same tensor $\cT$, then
\begin{equation}\label{eq:canonicalization_identity}
 \Can(\tn)=\Can(\tn'),\qquad
 \jointcontract_E\bigl(\Can(\tn)\bigr)=\cT .
\end{equation}
Equality of canonicalizations holds core by core, including bond dimensions.
\end{theorem}

The proof is in Appendix~\ref{sec:tt_canonical_proof}.

\subsection{TT inner products}\label{subsec:ttinnerproduct}
Dense evaluation becomes impractical as $N$ grows, so we compute
the Frobenius norms and inner products in the training loss directly
from the predicted and target TT cores. For real tensors $\cT$ and
$\widehat{\cT}$, the squared residual is
\begin{equation}\label{eq:tt_residual_norm}
 \big{\|}\widehat{\cT}-\cT\big{\|}_F^2
 = \big{\langle} \widehat{\cT}, \widehat{\cT} \big{\rangle}
 -2\big{\langle} \cT, \widehat{\cT}\big{\rangle}+\inner{\cT}{\cT}.
\end{equation}
For two TTs with matching open-mode dimensions $m_i$, write
$T_i(a)$ and $\widehat T_i(a)$ for their
core slices at open index $a$ and site $i$. The partial contraction $L_i$
over the first $i$ sites leaves the two right bonds free. It satisfies
\begin{equation}\label{eq:ttinnerproduct}
 L_0=[1],\qquad
 L_i=\sum_{a=1}^{m_i}T_i(a)^\top L_{i-1}\widehat T_i(a),\qquad
 \big{\langle} \cT,  \widehat{\cT}\big{\rangle}=L_N(1,1).
\end{equation}

At site $i+1$, the two free indices of $L_i$ contract with the left
bonds of the two cores there.
The TTIN losses in \cref{subsec:experiment_setting} all use this
contraction.

\subsection{TTIN's computational and memory costs}\label{subsec:operation_costs}
To form the predicted cores used in the loss, the TTIN backbone in
\cref{eq:tnin} applies $L$ local layers and $n$ t-attention layers on
$N$ sites. Let $r$ and $d$ bound the bond and
open-mode dimensions used in layer contractions, including those of
projected attention cores. For original open modes of size at most $m$,
we use $d=m$. With a shared orthonormal basis of $K\leq m$ vectors
in \cref{eq:fixedbasiscontraction}, the bounds for coefficient-space
contractions use $d=K$. Each site has a bounded number of local terms
spanning at most $q\leq 3$ cores. Each t-attention layer has $h$ heads
and a fixed number of output tuples per head.

We first analyze TTIN's memory cost. Each local intermediate has two
bond indices and at most $q$ open modes, giving $\mathcal{O}(Nr^2d^q)$ entries
per sample and layer. A t-attention layer uses
$\mathcal{O}\bigl(h(Nr^2d+N^2)\bigr)$ entries per sample for the projected cores
and all pair scores. At fixed $m,r,d,q,h$, training saves $\mathcal{O}(LN+nN^2)$
activation entries per sample. Inference reuses buffers across layers,
so peak activation storage per sample is $\mathcal{O}(N^2)$ when storing all
scores. The maps and bases use $\mathcal{O}(N)$ entries per layer, and each dense
t-attention bias has $hN^2$ entries. Including these resident
arrays gives a total memory bound of $\mathcal{O}(LN+nN^2)$ at fixed batch size.

Next, we analyze the forward and backward computation costs. Forming
the local terms in \cref{eq:ttlocalupdate} costs
$\mathcal{O}\bigl(Nr^2d^q(r+d)\bigr)$ operations per layer. In t-attention, the
projection and output maps act on each core, and the inner products
and weighted sums in \cref{eq:attention} process $N^2$ core pairs
per head. Each pair involves cores with at most $r^2d$ entries.
A shared orthonormal basis preserves the query--key inner products,
so we can compute them directly from the coefficients. The resulting cost is
$\mathcal{O}\bigl(hNr^2d(N+r+d)\bigr)$ per t-attention layer.
With the intermediates saved, backpropagation uses contractions of
comparable size. At fixed $m,r,d,q,h$, each local layer is linear in $N$
and each t-attention layer is quadratic, giving $\mathcal{O}(LN+nN^2)$
operations per backbone forward or backward pass.

\subsection{Experiment settings}\label{subsec:experiment_setting}
We test TTIN on three problems. HJB represents nonlinear evolution of
continuous fields. The Ising model illustrates a natural application to
quantum many-body problems, where Hamiltonians and states admit
tensor-network representations. Glauber dynamics represents evolution
with state constraints, since the state is a probability distribution over
discrete configurations. All three admit TT inputs and outputs, so we use
a common backbone layout.

Analytic encoders construct the input TTs.
The backbone first updates their cores through five local interaction
layers, using $\sigma=\tanh$ in \cref{eq:ttlocalupdate}, and then applies
t-attention.
Each head learns the projection tuples ${\mathbf W_{i,m}^{\mathrm{Q}}}$,
${\mathbf W_{i,m}^{\mathrm{K}}}$, and ${\mathbf W_{i,m}^{\mathrm{V}}}$ separately at
each core position and uses two output tuples in \cref{eq:attnoutput}.
With the predictions and targets in TT form, we use the inner products
in \cref{eq:ttinnerproduct} to compute the losses directly from the cores.

TTIN uses FP32 with AMP and TF32 disabled. We train on an NVIDIA GeForce
RTX 4060 Laptop GPU with 8\,GB of memory. The host has 96\,GB of memory
and an AMD Ryzen 9 7945HX CPU with 16 cores and 32 threads.
We use Adam with batch size $b=32$,
gradient norm clipping at one, zero weight decay, and a fixed epoch budget.
The learning rate rises linearly from
$10^{-3}$ to $10^{-2}$ over the first $5\%$ of epochs and then follows
cosine decay to $10^{-4}$. We evaluate the final checkpoint once on
the full training, validation, and test sets using the same batch size.
The accompanying GitHub demo illustrates the analytic input encoders
and core updates. The supplement gives further architecture details
and reference checks.

\subsubsection{\texorpdfstring{{HJB}}{HJB} evolution}
We learn the fixed-time solution map $u_0\mapsto u(T,\cdot)$ for the
periodic nonlinear equation
\begin{equation}\label{eq:hjb}
 \partial_tu=\nu\Delta u-\tfrac12|\nabla u|^2,\qquad
 u(0,\cdot)=u_0,\qquad x\in[0,2\pi)^N,\quad 0\le t\le T,
\end{equation}
with viscosity $\nu=0.2$ and final time $T=2$. We use $N=4,12,20,28,36$
and sample both fields on a tensor product grid with 16 points per
coordinate.
The initial field combines single-coordinate Fourier modes and
nearest-neighbor couplings:
\begin{align}\label{eq:hjb_initial}
 u_0(x)={}&\frac1{\sqrt N}\sum_{i=1}^N
     \left[a_i\cos(x_i-\theta_i)+c_i\cos(2x_i-\phi_i)\right]\nonumber\\
 &+\frac1{\sqrt{N-1}}\sum_{i=1}^{N-1}
      b_i\cos(x_i-x_{i+1}-\psi_i).
\end{align}
We draw $a_i\sim U(0.3,0.6)$, $c_i\sim U(0.05,0.15)$, and
$b_i\sim U(0.2,0.4)$ independently from uniform distributions, and take the phases
$\theta_i,\phi_i,\psi_i$ uniform on $[0,2\pi)$.

We compute TT reference solutions via Cole-Hopf \cite{hopf1950burgers}.
\Cref{tab:hjb_references} reports reliability checks for the reference
solutions.
Each size uses a 16,384/2,048/2,048
training/validation/test split.
An analytic TT of bond dimension four encodes the single-coordinate terms
and nearest-neighbor couplings. The encoder zero-pads the internal bonds
to dimension ten. In the local layers and t-attention, we apply
\cref{eq:fixedbasisweights} to the open modes using eight real Fourier
functions orthonormal under the uniform grid average
\cref{eq:hjb_fourier_basis}.

We train for 200 epochs at every size.
For the architecture comparison at $N=4$, we train with three different
random seeds and report the mean and sample standard deviation of the
test errors. We also average TTIN's training time over those seeds.
For the dimension sweep, we train once from scratch at each size using
the same fixed seed.
The loss is the mean relative Frobenius error of the terminal grid
tensor $\cT^{(j)}$,
\begin{equation}\label{eq:hjb_loss}
 \mathcal L_{\rm HJB}(\theta)=\frac1b\sum_{j=1}^{b}
 \frac{ \bigg{\|}\widehat{\cT}^{(j)}-\cT^{(j)}\bigg{\|}_F }
      {\norm{\cT^{(j)}}_F}.
\end{equation}

\begin{figure}[H]
\centering
\includegraphics[width=4.45in]{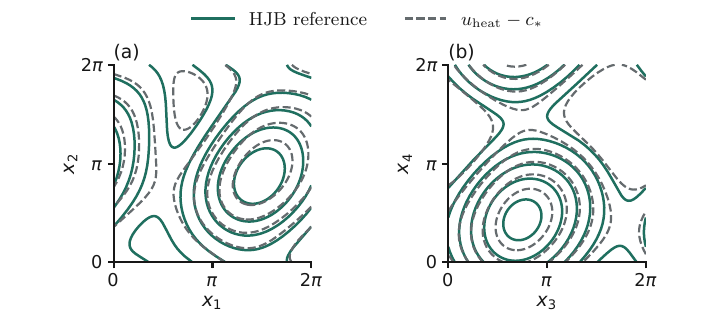}
\caption{HJB reference and shifted heat solution for the first $N=4$
validation input at $T=2$. Panel (a) varies $x_1,x_2$ with $x_3=x_4=0$.
Panel (b) varies $x_3,x_4$ with $x_1=x_2=0$. Both panels use the same
shift and contour levels, with periodic interpolation for display.}
\label{fig:hjb_nonlinearity}
\end{figure}

To quantify the effect of the nonlinear term in \cref{eq:hjb}, we compare
the HJB reference $u(T,\cdot)$ with the time-$T$ solution $u_{\mathrm{heat}}$
of $\partial_t v=\nu\Delta v$ with $v(0,\cdot)=u_0$.
On all 2,048 validation inputs at $N=4$, diffusion alone gives a mean
relative Frobenius error of $49.12\%$. To separate changes in spatial shape from
the mean shift, we subtract the grid average
$c_*=\E_x[u_{\mathrm{heat}}-u(T,\cdot)]$ from the heat solution for each
input. The mean relative error after this shift is $14.95\%$.
\Cref{fig:hjb_nonlinearity} shows the remaining differences in contour shapes
on two coordinate slices.

\subsubsection{Ising ground state prediction}
We learn $H\mapsto\psi_0$, with $\psi_0$ the normalized ground state of
the open-chain transverse-field Ising Hamiltonian
\begin{equation}\label{eq:ising}
\begin{gathered}
 H=-\sum_{i=1}^{N-1}J_i Z_iZ_{i+1}-\sum_{i=1}^{N}h_i X_i,\\
 Z_i=I_2^{\otimes(i-1)}\otimes
      \begin{pmatrix}1&0\\0&-1\end{pmatrix}\otimes I_2^{\otimes(N-i)},\quad
 X_i=I_2^{\otimes(i-1)}\otimes
      \begin{pmatrix}0&1\\1&0\end{pmatrix}\otimes I_2^{\otimes(N-i)},
\end{gathered}
\end{equation}
where $I_2$ is the identity on a single spin and $Z_i$ and $X_i$ are
the Pauli operators on site $i$. The coupling $J_i>0$
links spins $i$ and $i+1$, and $h_i>0$ is the transverse-field strength
at site $i$.

We use $N=16,32,64,128,256$. The base distribution uses independent
$J_i,\eta_i,\bar h\sim U(0.5,1.5)$ and sets $h_i=\bar h\eta_i$.
Near the transition between coupling-dominated and field-dominated
behavior, small changes in $J_i$ or $h_i$ can substantially change the
ground state. We measure the balance between fields and couplings by
\begin{equation}\label{eq:ising_distance}
 d_N=\frac1N\sum_{i=1}^N\log h_i
       -\frac1{N-1}\sum_{i=1}^{N-1}\log J_i.
\end{equation}
We locate this region near $d_N=0$, where the geometric means of the
fields and couplings agree. To sample more densely on the lower-field
side, each training set combines 1,024 inputs from the base distribution
with 1,024 additional inputs concentrated around $d_N=-0.1$.
The validation and test sets each contain 256 inputs from the base
distribution. We use the free-fermion reduction \cite{pfeuty1970ising}
to obtain the Gaussian
reference states and convert them to MPSs of
local dimension two \cite{fishman2015gaussian}.
\Cref{tab:ising_references} reports reliability checks for the
reference solutions.

We encode the local fields and neighboring couplings as an analytic MPO
of bond dimension three. The encoder combines the two physical indices
of each core and zero-pads the internal bonds to dimension eight.
The weight maps act in the original physical coordinates.
The decoder uses modewise linear maps and a bias to produce MPS cores.
It rescales each core by its Frobenius norm to stabilize the contractions
in the loss. The overall MPS scale cancels from \cref{eq:fidelity}.
We train for 200 epochs at each size using the same fixed seed.
The loss is mean infidelity
$b^{-1}\sum_{j=1}^{b}[1-F(\widehat\psi^{(j)},\psi_0^{(j)})]$,
where $\widehat\psi$ is the decoded MPS and
\begin{equation}\label{eq:fidelity}
 F(\widehat\psi,\psi_0)=
 \frac{\Bigl|\,\inner{\widehat\psi}{\psi_0}\,\Bigr|^2}
      {\Bigl\lVert\widehat\psi\Bigr\rVert^2\,\Bigl\lVert\psi_0\Bigr\rVert^2}.
\end{equation}

\subsubsection{Glauber probability evolution}
Given an initial distribution $p_0$, we predict $p(T)=e^{TQ}p_0$ on
an open chain of $N$ spins $s_i\in\{-1,1\}$. The off-diagonal transitions
of the continuous-time Markov generator $Q$ are single-spin flips.
In configuration $s$, the rate for flipping spin $i$ is
\begin{equation}\label{eq:glauber}
 c_i(s)=\tfrac12\left[1-s_i\tanh\!\left(B+K\sum_{j\sim i}s_j\right)\right],
\end{equation}
where the sum is over the neighbors of $i$. The diagonal entries make
each column of $Q$ sum to zero. We use nearest-neighbor coupling $K=0.8$,
external field $B=0.4$, and final time $T=0.5$ on chains of 20, 40, 60,
80, and 100 spins. Each input is a mixture
of two product distributions,
\begin{equation}\label{eq:glauber_initial}
 p_0(s)=w\prod_i\frac{1+m_i^+s_i}{2}
      +(1-w)\prod_i\frac{1+m_i^-s_i}{2},
\end{equation}
where $w\sim U(0.3,0.7)$, $m_i^+\sim U(0.2,0.7)$, and
$m_i^-\sim U(-0.7,-0.2)$ independently.

Each size uses a 2,048/256/256 training/validation/test split.
We generate the targets with {TEBD} \cite{vidal2004tebd}. \Cref{tab:glauber_references}
reports reliability checks for the reference solutions.
An analytic TT uses two bond channels to encode the two product
distributions. The encoder zero-pads the internal bonds to dimension
six. Each local layer uses the single-core term in
\cref{eq:ttlocalupdate}, and the weight maps act in the original spin
coordinates.

We train one model per size for 2,000 epochs using the same fixed seed.
In the increasing-size training sequence, each warm start takes the
backbone and t-attention weights from the preceding run's final
checkpoint. We restart the optimizer, learning-rate schedule, and
sampler at each size.
The loss is the mean relative $L^2$ error of the probability vector,
\begin{equation}\label{eq:glauber_loss}
 \mathcal L_{\rm GL}(\theta)=\frac1b\sum_{j=1}^{b}
 \frac{\norm{\widehat p^{(j)}(T)-p^{(j)}(T)}_2}
      {\norm{p^{(j)}(T)}_2}.
\end{equation}

\Needspace{4\baselineskip}
\subsubsection{Effect of canonicalization on training and prediction}
\label{subsec:canonicalization_study}
To assess how gauge symmetry affects test accuracy, we train separate
TTINs for each problem on analytic cores, their canonicalized forms,
and raw TT-SVD cores.
For all three problems, we construct the canonicalized inputs with
\cref{alg:tt_canonicalization}, using the same $\operatorname{FixSVD}$
rule in Appendix~\ref{sec:tt_canonical_proof}.

\begin{table}[htbp]
\centering
\small
\caption{TTIN test errors (\%) for three input representations.}
\label{tab:canonicalization_comparison}
\setlength{\tabcolsep}{4pt}
\renewcommand{\arraystretch}{1.12}
\begin{tabular*}{\linewidth}{@{\extracolsep{\fill}}lrrr@{}}
\toprule
Problem & Analytic encoding & Canonicalized & Raw TT-SVD \\
\midrule
HJB ($N=4$) & 0.1799 & 0.2972 & 0.6950 \\
Glauber ($N=20$) & 0.0977 & 0.1518 & 0.2240 \\
Ising ($N=16$) & 0.0074 & 0.8264 & 15.0314 \\
\bottomrule
\end{tabular*}
\end{table}

\Cref{tab:canonicalization_comparison} shows that canonicalized inputs
give lower errors than raw TT-SVD in all three problems. Analytic
encodings give the lowest errors in each case. This suggests that
placing known components of the inputs in fixed core coordinates
across samples makes the map easier to learn.

\Needspace{5\baselineskip}
To assess how gauge symmetry affects predictions, we compare the
outputs of the same trained TTIN for each canonicalized test TT and
its gauge-equivalent counterpart. We obtain the counterpart by
flipping the sign of the first coordinate on the middle bond via
\cref{eq:gauge}. We repeat the comparison after canonicalizing both
inputs with \cref{alg:tt_canonicalization}.

\begin{table}[H]
\centering
\small
\caption{Mean prediction differences (\%) for gauge-equivalent inputs.}
\label{tab:gauge_consistency}
\setlength{\tabcolsep}{4pt}
\renewcommand{\arraystretch}{1.12}
\begin{tabular*}{\linewidth}{@{\extracolsep{\fill}}lrr@{}}
\toprule
Problem & Prediction difference & After canonicalization \\
\midrule
HJB ($N=4$) & 14.21 & 0 \\
Glauber ($N=20$) & 13.39 & 0 \\
Ising ($N=16$) & 48.85 & 0 \\
\bottomrule
\end{tabular*}
\end{table}

Changing the gauge substantially alters TTIN predictions
(\cref{tab:gauge_consistency}). Canonicalizing both inputs first
eliminates the differences for all tested pairs, giving consistent
predictions when an analytic encoding is hard to construct.

\Needspace{5\baselineskip}
Taken together, the two experiments favor a common analytic encoding
whenever the problem provides enough structure to construct one.
Such an encoding makes that structure explicit in the cores, which can
make the map easier to learn. When a common encoding is difficult to
construct, we apply \cref{alg:tt_canonicalization} to training and
prediction inputs to improve test accuracy and obtain consistent
predictions across equivalent representations.

\Needspace{6\baselineskip}
\section{Numerical Results}\label{sec:experiments}
To assess the value of direct core interactions, we compare TTIN with
models that vectorize the same input cores. We then examine how accuracy
and cost scale with problem size and measure the end-to-end speedups of
TTIN-based pipelines over established numerical solvers at comparable
accuracy. We also test the construction beyond operator learning with
a TT-CNN for human activity recognition.

To compare runtimes on the operator learning problems, we measure
end-to-end latency from inputs in CPU memory to final outputs in CPU
memory for batches of one (B1) and 32 (B32).
To estimate when the latency savings offset TTIN's training time $C$,
we report the smallest batch count $k$ such that
$C+k\tau_b^{\rm TTIN}<k\tau_b^{\rm solver}$, where
$\tau_b^{\rm TTIN}$ and $\tau_b^{\rm solver}$ are the measured
latencies at batch size $b$.

\subsection{\texorpdfstring{{HJB}}{HJB} evolution}\label{sec:hjb}
We first compare TTIN at $N=4$ with vectorized-core models and
operator learning methods of comparable parameter counts in terms of
test error, training time, and inference latency (\cref{tab:hjb_comparison}).
We then examine how TTIN scales with $N$ and estimate its break-even
batch count relative to a TT Cole-Hopf solver.

\subsubsection{Comparison with other learning methods}
To examine the role of direct interactions between tensor cores, we
vectorize the same input cores and concatenate them into a single vector
for the residual {MLP} or arrange them as a sequence
for the Transformer and {1D} CNN. We reshape the outputs to TT core dimensions. The other
settings follow TTIN's.

Unless otherwise stated, the neural network baselines follow the shared
settings in \cref{subsec:experiment_setting}.
We train the {FNO}~\cite{li2021fno} using
NeuralOperator~\cite{kossaifi2025neuraloperatorlibrary} for 20 epochs
with batch size four, counting each complex spectral coefficient as
two real parameters. For {POD-DeepONet}~\cite{lu2022poddeeponet}, we use
the DeepXDE implementation \cite{lu2021deepxde} and align phases to
reduce variation from translations before fitting the POD basis. We use a POD
dimension of 192 selected on validation data and train the branch with
one-tenth of the shared learning rates.
{TT-ROM} adapts THOR
\cite{danis2025ttopinf} to fit quadratic reduced dynamics on separate
trajectories with a comparable number of training fields. We use the full
validation set to select FNO checkpoints and TT-ROM regularization
parameters. TT-ROM's offline time includes validation and test rollouts.

\begin{table}[!htbp]
\centering
\small
\caption{HJB architecture comparison at $N=4$.}\label{tab:hjb_comparison}
\setlength{\tabcolsep}{4pt}
\renewcommand{\arraystretch}{1.12}
\begin{tabular*}{\linewidth}{@{\extracolsep{\fill}}lrrrrr@{}}
\toprule
Method & Parameters & Test error (\%) & Training (s) & B1 (ms) & B32 (ms) \\
\midrule
TTIN & 491,132 & $0.1806\pm 0.0007$ & 355 & 0.662 & 1.97 \\
\midrule
Residual MLP & 488,914 & $7.291\pm 1.151$ & 65.7 & 0.173 & 1.18 \\
Core Transformer & 490,890 & $0.742\pm 0.035$ & 93.9 & 0.190 & 1.20 \\
Core 1D CNN & 490,561 & $1.427\pm 0.215$ & 95.1 & 0.241 & 1.26 \\
\midrule
FNO & 491,030 & $0.0929\pm 0.0015$ & 1,369 & 1.21 & 94.5 \\
POD-DeepONet & 491,126 & $0.5512\pm 0.0017$ & 208 & 1.57 & 2.75 \\
TT-ROM & 491,328 & $1.7133\pm 0.0017$ & 309 & 9.19 & 267 \\
\bottomrule
\end{tabular*}
\end{table}

\Needspace{3\baselineskip}
At similar parameter counts, TTIN achieves about one quarter of the
lowest mean test error among the three baselines that vectorize the
same input cores. This gap suggests that vectorization obscures the
structure encoded by the TT bonds and thereby makes the evolution
harder to learn. It illustrates the benefit of preserving this
structure through direct interactions between tensor cores.

Among the operator learning methods, TTIN has lower test error and
inference latency than POD-DeepONet and TT-ROM. POD-DeepONet trains faster than
TTIN. FNO has the lowest test error, and TTIN has shorter training time
and lower inference latency than FNO at both batch sizes.
Together, these results show that TTIN combines high accuracy with
computational efficiency. For problems whose inputs and outputs admit
compact TT representations, TTIN can serve as a practical alternative
to established operator learning methods.

\Needspace{6\baselineskip}
\subsubsection{Scalability and cost of repeated solves}
{In subsection~5.1.1, we compared TTIN at $N=4$ with
vectorized baselines, FNO, POD-DeepONet, and TT-ROM. At higher dimensions,
the full $16^N$-point grid makes the FNO and POD-DeepONet implementations
impractical to run, and THOR's current TT-ROM implementation does not support 
larger HJB problem sizes as considered here. In experiments at larger $N$ (not shown),
the vectorized baselines yielded substantially higher mean test errors
than TTIN. Therefore, we examine how TTIN scales over $N=4,12,20,28,36$ and
compare it with TT Cole-Hopf at comparable accuracy to assess the cost of
repeated solves.}

\begin{table}[!htbp]
\centering
\small
\caption{HJB prediction and Cole-Hopf with \texttt{tntorch}.}\label{tab:hjb_vertical}
\setlength{\tabcolsep}{4pt}
\renewcommand{\arraystretch}{1.12}
\begin{tabular*}{\linewidth}{@{\extracolsep{\fill}}llrrr@{}}
\toprule
$N$ & Method & Test error (\%) & B1 (ms) & B32 (ms) \\
\midrule
4 & TTIN & 0.1838 & 0.690 & 1.87 \\
4 & Cole-Hopf + \texttt{tntorch} & 0.1574 & 9.61 & 77.8 \\
\midrule
12 & TTIN & 0.0764 & 0.906 & 4.50 \\
12 & Cole-Hopf + \texttt{tntorch} & 0.0772 & 25.1 & 144 \\
\midrule
20 & TTIN & 0.0618 & 1.22 & 7.73 \\
20 & Cole-Hopf + \texttt{tntorch} & 0.0531 & 44.4 & 228 \\
\midrule
28 & TTIN & 0.0664 & 1.47 & 11.7 \\
28 & Cole-Hopf + \texttt{tntorch} & 0.0714 & 85.0 & 545 \\
\midrule
36 & TTIN & 0.0765 & 1.57 & 16.0 \\
36 & Cole-Hopf + \texttt{tntorch} & 0.0712 & 103 & 716 \\
\bottomrule
\end{tabular*}
\end{table}

\Needspace{2\baselineskip}
The mean test error drops from $N=4$ to $N=12$ and remains below
$0.08\%$ at the larger sizes (\cref{tab:hjb_vertical}).
In increasing order of $N$, training updates take 370, 1,206, 1,785, 2,608,
and 3,534\,s. The update time grows approximately
linearly with $N$. The parameter count rises from 491,132 at $N=4$ to
1,535,612 at $N=36$, about a threefold increase as $N$ grows ninefold.
These trends suggest that TTIN scales well across dimensions.

\Needspace{3\baselineskip}
We next compare the median inference latencies of TTIN and a TT Cole-Hopf
solver at similar mean test errors. The solver evolves the transformed
field in TT form and uses TT-cross with \texttt{tntorch}~\cite{usvyatsov2022tntorch}
to approximate the inverse transform.
Both methods' measured latencies increase with $N$
(\cref{tab:hjb_vertical}). TTIN has lower latency at every size, with
speedups of about $65\times$ for B1 and $45\times$ for B32 at $N=36$.
In increasing order of $N$, the break-even counts are 41,433, 49,921,
41,324, 31,214, and 34,967 for B1, and 4,870, 8,630, 8,112, 4,887,
and 5,052 for B32.
For applications with moderate to high demand for repeated solves,
TTIN offers a clear cost advantage.
\suppressfloats[t]
\begin{table}[H]
\centering
\small
\caption{Mean test fidelity (\%) of direct TTIN predictions and their refinements.}\label{tab:ising_stages}
\setlength{\tabcolsep}{4pt}
\renewcommand{\arraystretch}{1.12}
\begin{tabular*}{\linewidth}{@{\extracolsep{\fill}}lrrrr@{}}
\toprule
$N$ & TTIN & TTIN + $\mathcal{P}$ & TTIN + DMRG & TTIN + DMRG + $\mathcal{P}$ \\
\midrule
16 & 99.992573 & 99.994529 & 99.832651 & 99.999994 \\
32 & 99.966120 & 99.972410 & 99.387143 & 99.999962 \\
64 & 99.875202 & 99.891337 & 97.985473 & 99.999189 \\
128 & 99.529412 & 99.562937 & 94.236788 & 99.975437 \\
256 & 98.355774 & 98.420637 & 82.810382 & 99.498575 \\
\bottomrule
\end{tabular*}
\end{table}

\subsection{Ising ground state prediction}\label{sec:ising}
To assess the scalability of TTIN for Ising ground state prediction,
we test it on chains with $N=16,32,64,128,256$. The mean test fidelity
of the direct predictions gradually decreases as the chains grow longer
(\cref{tab:ising_stages}). To improve fidelity on longer chains, we test
two refinement steps after the learned readout.
The first step uses YASTN's two-site DMRG \cite{rams2025yastn}.
It lowers the energy of the predicted MPS by sweeping along
the chain and optimizing adjacent pairs of cores.
We next apply even-parity projection because the ground state $\psi_0$ in
\cref{eq:ising} lies entirely in the even-parity sector of the global spin
flip $P=X_1\cdots X_N$ \cite{campanino2023inhomogeneous}.
TTIN predictions can contain odd-parity components, so projecting onto
the even sector may improve fidelity. We write $\mathcal{P}$ for the
map that applies this projection, rounds the result to bond dimension
at most eight, and normalizes it.

Applying DMRG alone to the TTIN predictions lowers their fidelity.
During the sweeps, odd-parity components can gain weight and reduce
overlap with the even ground state. At $N=256$, for example, the mean
odd-parity weight rises from $0.067\%$ to $16.8\%$. Projection with $\mathcal{P}$ alone
gives a small fidelity gain. DMRG followed by $\mathcal{P}$ gives the
highest mean fidelity at each size (\cref{tab:ising_stages}), so we
use this sequence in the decoder.
In increasing order of $N$, TTIN's training times are 189, 349, 690,
1,410, and 3,103\,s. These times are roughly proportional to $N$.

To measure the value of initializing DMRG with the TTIN readout, we
compare with DMRG started from a fixed product state and followed by
$\mathcal{P}$.
At similar test fidelity, we compare mean relative energy error
and end-to-end latency
(\cref{tab:ising_inference}). For reference energy
$\varepsilon_0$ and computed energy $\varepsilon$, the reported energy
error is $100\,\E[|\varepsilon-\varepsilon_0|/|\varepsilon_0|]$.
Both pipelines give similarly small errors at each size.

\begin{table}[H]
\centering
\small
\caption{Mean test fidelity, mean relative energy error, and end-to-end latency for Ising.}\label{tab:ising_inference}
\setlength{\tabcolsep}{4pt}
\renewcommand{\arraystretch}{1.12}
\begin{tabular*}{\linewidth}{@{\extracolsep{\fill}}llrrrr@{}}
\toprule
$N$ & Method & $F$ (\%) & Energy error (\%) & B1 (s) & B32 (s) \\
\midrule
16 & DMRG + $\mathcal{P}$ & 99.999991 & $2.22\times10^{-5}$ & 0.320 & 2.41 \\
16 & TTIN + DMRG + $\mathcal{P}$ & 99.999994 & $2.09\times10^{-5}$ & 0.0597 & 0.239 \\
\midrule
32 & DMRG + $\mathcal{P}$ & 99.999960 & $2.82\times10^{-5}$ & 1.52 & 11.7 \\
32 & TTIN + DMRG + $\mathcal{P}$ & 99.999962 & $2.83\times10^{-5}$ & 0.119 & 0.484 \\
\midrule
64 & DMRG + $\mathcal{P}$ & 99.993848 & $5.61\times10^{-5}$ & 5.18 & 43.3 \\
64 & TTIN + DMRG + $\mathcal{P}$ & 99.999189 & $4.75\times10^{-5}$ & 0.241 & 0.963 \\
\midrule
128 & DMRG + $\mathcal{P}$ & 99.976232 & $1.08\times10^{-4}$ & 34.0 & 284 \\
128 & TTIN + DMRG + $\mathcal{P}$ & 99.975437 & $8.52\times10^{-5}$ & 0.453 & 1.94 \\
\midrule
256 & DMRG + $\mathcal{P}$ & 99.509250 & $1.93\times10^{-4}$ & 85.0 & 706 \\
256 & TTIN + DMRG + $\mathcal{P}$ & 99.498575 & $2.30\times10^{-4}$ & 0.968 & 3.82 \\
\bottomrule
\end{tabular*}
\end{table}

The TTIN pipeline's latency grows approximately linearly with $N$
(\cref{tab:ising_inference}). Its speedup over the pure DMRG pipeline
increases with $N$ at both batch sizes, reaching about $88\times$ for B1
and $185\times$ for B32 at $N=256$. In increasing order of $N$, the
break-even counts are 727, 250, 140, 43, and 37 for B1, and 87, 32, 17,
5, and 5 for B32. These results suggest that the TTIN pipeline can be
highly cost-effective for repeated ground-state solves in quantum
many-body problems.
\subsection{Glauber probability evolution}\label{sec:glauber}
To assess TTIN's scalability for probability evolution, we test it
on Glauber chains with $N=20,40,60,80,100$. As the configuration space
grows from $2^{20}$ to $2^{100}$ states, TTIN's mean test error
increases gradually by less than a factor of two and stays low
across all five chain lengths.

To compare runtimes at similar accuracy, we benchmark TTIN against
quimb's TEBD \cite{gray2018quimb}. With a fivefold increase in chain
length, TTIN's latency grows by factors of about $1.9$ for B1 and $2.4$
for B32. The corresponding factors for the selected TEBD configurations
are about $12$ and $15$. TTIN has lower latency at every size, with
speedups reaching about $259\times$ for B1 and $2{,}176\times$ for B32
at $N=100$ (\cref{tab:glauber}).
The speed advantage is more pronounced on longer chains.

We also check how closely the predictions satisfy the probability
constraints. We report the mean mass deviation
$\delta_{\mathrm{mass}}=\E[\lvert\sum_s\widehat p(s)-1\rvert]$
and the fraction $f_{\geq 0}$ of checked entries with $\widehat p(s)\geq 0$.
Nearly all checked TTIN entries are nonnegative. TTIN's mean mass
deviation increases with $N$ and exceeds TEBD's at each size.
Overall, TTIN's predictions adhere closely to both probability constraints.

In increasing order of $N$, TTIN's training times are 212, 410, 967,
1,327, and 1,764\,s. We sum the training times along the full warm-start
chain through each size to compute the break-even counts. In the same
order, the break-even counts are 25,006, 15,924, 26,806, 25,898, and
29,809 for B1, and 1,841, 1,567, 2,706, 1,895, and 1,856 for B32.
The per-batch savings make TTIN a cost effective option for
applications that repeatedly predict the evolution of constrained
states in large batches.

\begin{table}[H]
\centering
\small
\caption{Glauber prediction and quimb TEBD.}\label{tab:glauber}
\setlength{\tabcolsep}{3pt}
\renewcommand{\arraystretch}{1.12}
\begin{tabular*}{\linewidth}{@{\extracolsep{\fill}}llrrrrr@{}}
\toprule
$N$ & Method & Test error (\%) & $\delta_{\mathrm{mass}}$ (\%) & $f_{\geq 0}$ (\%) & B1 (ms) & B32 (ms) \\
\midrule
20 & TTIN & 0.0828 & 0.0369 & 100.0000 & 0.345 & 0.525 \\
20 & TEBD & 0.0770 & 0.0260 & 100.0000 & 14.1 & 187 \\
\midrule
40 & TTIN & 0.1111 & 0.0980 & 100.0000 & 0.386 & 0.679 \\
40 & TEBD & 0.1313 & 0.0031 & 100.0000 & 66.9 & 677 \\
\midrule
60 & TTIN & 0.1205 & 0.1649 & 99.9908 & 0.546 & 0.913 \\
60 & TEBD & 0.1183 & 0.1363 & 100.0000 & 76.1 & 750 \\
\midrule
80 & TTIN & 0.1233 & 0.2962 & 99.8878 & 0.615 & 1.12 \\
80 & TEBD & 0.1529 & 0.0204 & 100.0000 & 130 & 1,771 \\
\midrule
100 & TTIN & 0.1555 & 0.6764 & 99.5087 & 0.665 & 1.27 \\
100 & TEBD & 0.1892 & 0.0280 & 99.9992 & 172 & 2,760 \\
\bottomrule
\end{tabular*}
\end{table}
\subsection{Human activity recognition on DSADS}\label{sec:dsads}
To test the versatility of our tensor-network construction beyond
operator learning, we apply TT-CNN to cross-subject activity recognition
on {DSADS}~\cite{barshan2010dsads}.
The recordings contain time series indexed by body location, sensor
modality, and spatial direction. To preserve these measurement axes,
we form one core per short time block, with one mode for each axis
and a fourth for time within the block. Shared temporal filters then
combine neighboring cores along the resulting sequence.

On HAROOD's four cross-subject folds~\cite{lu2026harood}, we compare
TT-CNN with a conventional 1D CNN of comparable parameter count.
We use the same training settings for both models and select checkpoints
based on source-domain validation accuracy.
\Cref{tab:dsads} also includes uniform risk minimization (URM),
HAROOD's strongest reported method under the same validation setting.

TT-CNN achieves {SOTA} mean test accuracy of $91.35\%$
on HAROOD's DSADS cross-subject splits under source-domain validation.
It surpasses URM by $2.34$ percentage points with about $22\%$ of URM's
estimated parameter count. It also improves on the comparably sized
1D CNN by $7.53$ percentage points, with higher accuracy in all four
folds (\cref{tab:dsads_folds}).
The improvement on held-out subjects shows how corewise convolution
can support accurate classification of multichannel time series with
structured measurements. Together with the HJB architecture comparison,
this result suggests that direct interactions among tensor cores can
improve accuracy on a broader range of tasks with structured inputs.
\Cref{sec:dsads_details} gives the convolution and training details.

\begin{table}[H]
\centering
\small
\caption{Cross-subject activity recognition on DSADS.}\label{tab:dsads}
\setlength{\tabcolsep}{4pt}
\renewcommand{\arraystretch}{1.12}
\begin{tabular*}{\linewidth}{@{\extracolsep{\fill}}lrr@{}}
\toprule
Method & Parameters & Test accuracy (\%) \\
\midrule
URM (Transformer) & $\approx 625{,}000$ & 89.01 \\
1D CNN & 139,187 & 83.82 \\
TT-CNN & 137,644 & 91.35 \\
\bottomrule
\end{tabular*}
\end{table}
\section{Discussion}\label{sec:discussion}
In this paper we  introduced a unified  architecture 
for learning maps between high-dimensional spaces via tensor network structured neural networks.  Our backbone updates tensor
cores through direct nonlinear interactions, keeping intermediate states in compact tensor network form. 
Task-specific encoders and decoders adapt the construction to different inputs and 
outputs. In numerical experiments, direct core interactions improved accuracy, and compact
representations supported accurate prediction in high dimensional state spaces. In high-dimensional operator
learning, our TTIN architecture substantially shortened solve times by predicting solutions directly or by providing 
initial states for a numerical solver in multiple problems. We also showed that the framework  supports classification 
of structured time series, as demonstrated by TT-CNN.
The results suggest that tensor network structured neural networks
can be powerful tools in high-dimensional applications.

To extend these uses to new problems, we could learn the underlying graph structures more directly or better adapt bond dimensions to
reduce storage. 
Sharing update parameters across
positions could also let one trained model handle different chain
lengths. When reference solutions are costly to generate, objectives
based on equation residuals or variational energies, such as Grimm and
Lindsey's energy loss~\cite{grimm2026foundation}, could reduce the need
for reference data. Testing the construction on a wider range of tasks
would help identify the relationships in the inputs and outputs that matter most for
accurate and efficient learning.  

\begingroup
\appendix
\Needspace{5\baselineskip}
\section{TT Canonicalization}
\label{sec:tt_canonical_proof}

\paragraph{Choice of FixSVD}
For a nonzero unfolding $T_i$, take a compact {SVD}
$T_i=U_i\Sigma_i V_i^\top$ retaining the positive singular values.
Let $\mathcal{G}$ apply Gram--Schmidt in order, skipping zero
residuals and dividing each nonzero residual by its norm. For the
coordinate vectors $e_1,\ldots,e_{m_i r_i}$ in the column order of
$T_i$, the isometry $V_i$ gives
\begin{equation}\label{eq:fixsvd_gram_schmidt}
 W_i=\operatorname{FixSVD}(T_i)
 =V_i\mathcal{G}\bigl((V_i^\top e_j)_j\bigr)
 =\mathcal{G}\bigl((V_iV_i^\top e_j)_j\bigr).
\end{equation}
The projector $V_iV_i^\top$ is determined by $T_i^\top T_i$, and
the projected coordinate vectors span the row space of $T_i$. Thus
the last expression gives a fixed orthonormal basis of that space.
The middle expression computes it in the smaller SVD coordinates.

\paragraph{Proof of \cref{thm:tt_canonical}}
For $\cT=0$, the convention fixes the output. For $N=1$, the sole
core is $\cT$. Otherwise, the forward QR sweep preserves $\cT$
by absorbing each $R_i$ into the next core. For $i<N$, let $P_i$
contract the first $i$ QR cores, with the open modes in the rows
and the right bond in the columns. With $P_0=[1]$ and $Q_i^\top Q_i=I$,
the recurrence gives column orthonormality by induction.
\begin{equation}\label{eq:tt_prefix_isometry}
 P_i=(P_{i-1}\otimes I_{m_i})Q_i,\qquad P_i^\top P_i=I .
\end{equation}

For the reverse sweep, assume inductively that both runs have the
same remaining tensor and saved suffix cores. At $i=N$, the remaining
tensor is $\cT$ and the suffix is empty. At site $i$, unfold the
remaining tensor as $M_i$, with $j_1,\ldots,j_{i-1}$ indexing rows
and $j_i$ and the exposed right bond indexing columns. The endpoint
unfolding $T_i$ satisfies
\begin{equation}\label{eq:tt_remaining_gram}
 M_i=P_{i-1}T_i,\qquad T_i^\top T_i=M_i^\top M_i .
\end{equation}
The common Gram matrix determines the same $W_i$ in both runs,
so the reshaped $W_i^\top$ gives the same saved core and bond dimension.
The mode-3 update replaces $Q_{i-1}$ by $Q_{i-1}(T_iW_i)$, so
the updated tensor unfolds to $P_{i-1}(T_iW_i)=M_iW_i$.
Moving $j_{i-1}$ into the columns with the new bond gives a common
$M_{i-1}$. Since the columns of $W_i$ span the row space of $M_i$,
\begin{equation}\label{eq:tt_row_space}
 M_i=(M_iW_i)W_i^\top .
\end{equation}
Thus $M_iW_i\ne 0$, and reattaching the saved core recovers $M_i$.
At the end of the sweep, both runs also have the same first core.
Reattaching the saved cores recovers $\cT$.\nobreak\hfill\proofbox

\section*{Acknowledgements} 
J.K. is grateful to Yifan Zhang for several helpful conversations, including in brainstorming phases of the project a year and a half ago.  
J.K. also thanks Nick Nelsen for suggesting to use the Tucker factorization \eqref{eq:fixedbasisweights} with fixed basis matrices to achieve discretization-independence for the number of learnable parameters.  
This research was supported in part by NSF DMS
2309782, NSF DMS 2436499, NSF CISE-IIS 2312746, DE SC0025312, the Sloan Foundation, and a Moncrief Grand Challenge Faculty Award from UT Austin. 

\section*{Statement on AI use}
Key ideas in this work are human contributions: developing neural architectures which layerwise  are tensor networks and which act on tensor cores directly; canonicalizing input tensor networks so as to ensure independence of the choice of gauge symmetries in the input; enforcing locality in the interaction between cores in the same layer (for both physical and computational reasons); tensorizing various popular standard neural architectures to obtain tensor network versions; and developing applications to high-dimensional operator learning.  At the same time, AI tools helped with the work from near its beginning.  The authors used GPT-6 to assist with drafting, literature organization, coding,  preparation of figures and tables, and checks of the manuscript. The authors assume responsibility for all content.

\section*{Code availability}
The accompanying \href{https://github.com/Ruize-C/TENSOR-NETWORK-STRUCTURED-NEURAL-NETWORKS}{GitHub repository} provides
compact implementations of TTIN,
TT-CNN, and TT canonicalization. It includes runnable examples for HJB, Ising,
Glauber, and DSADS, with reference data generators for the three operator learning
examples and instructions for obtaining the DSADS data.

\bibliographystyle{siamplain}
\Needspace{5\baselineskip}
\enlargethispage{2\baselineskip}
\vspace{-4pt}
\bibliography{references}

\clearpage
\endgroup

\begingroup
\renewcommand{\siamprelabel}{SM}
\setcounter{section}{0}
\setcounter{subsection}{0}
\setcounter{page}{1}
\renewcommand{\itdefault}{\updefault}
\headers{Supplementary Materials}{R. Huang and J. Kileel}
\begin{center}
  \normalfont\large\bfseries Supplementary Materials
\end{center}

We document the tensor train (TT) interaction network (TTIN) experiments,
extend \cref{alg:tt_canonicalization} to trees, and describe the tensor train CNN (TT-CNN) experiment on the Daily and Sports
Activities dataset (DSADS).

\section{TTIN Implementation Details}\label{sec:other_architectures}
\subsection{Local interactions at the chain ends}\label{subsec:local_boundaries}
For $N\geq4$, let $\cT_i$ be core $i$ at a fixed layer, with bond
dimensions as in \cref{eq:tt}. Set
$\alpha_{i,2}^-=\alpha_{i,2}^+=0$ at both ends in \cref{eq:ttlocalupdate}.
At site~1, the term with coefficient $\alpha_{1,3}$ has the form
\begin{equation}\label{eq:tt_boundary_first}
 \alpha_{1,3}\bigl[\mathcal W_{1,3}\contract
    (\cT_1\contract\cT_2)\bigr]\times_1 L_{1,3}\times_3 R_{1,3},
\end{equation}
where $L_{1,3}$ is scalar and $R_{1,3}\in\mathbb R^{r_1\times r_2}$.
At site~2, the spans $(1,2)$ and $(1,2,3)$ reach the left end.
Write $\mathcal X$ for either contracted span with its singleton left
bond removed. The weight $\mathcal W$ produces the left bond of size
$r_1$ jointly with the output open mode and maps the exterior right bond
to size $r_2$ if needed. With the modes of $\mathcal S=\mathcal W\contract
\mathcal X$ ordered as left bond, open mode, and right bond, the
contribution with coefficient $\alpha$ is
\begin{equation}\label{eq:tt_boundary_second}
 \alpha\,\mathcal S\times_1 L\times_3 R,\qquad
 L\in\mathbb R^{r_1\times r_1},\qquad R\in\mathbb R^{r_2\times r_2}.
\end{equation}
The $\alpha_{2,2}^+$ term uses cores 2 and 3. Reversing the site order
and exchanging the roles of the two bonds gives the right-end rules. In the Hamilton-Jacobi-Bellman (HJB) and
Ising experiments, each term involving multiple cores has its own learned
$L$ and $R$ at each site and layer.

\subsection{Fixed-basis contractions in HJB}\label{subsec:basis_analysis}
On the grid $x_j=2\pi j/m$, $j=0,\ldots,m-1$, with $m=16$, we use
the real Fourier functions $(\varphi_1,\ldots,\varphi_8)$ given by
\begin{equation}\label{eq:hjb_fourier_basis}
 \bigl(1,\,\sqrt{2}\cos x,\,\sqrt{2}\sin x,\,
 \sqrt{2}\cos 2x,\,\sqrt{2}\sin 2x,\,
 \sqrt{2}\cos 3x,\,\sqrt{2}\sin 3x,\,\sqrt{2}\cos 4x\bigr).
\end{equation}
The grid evaluation matrix is $\sqrt m\,B$, where $B$ has orthonormal
columns as in
\cref{eq:fixedbasiscontraction}. We use the uniform grid inner
product with weight $1/m$ at each point. The adjoint of the evaluation
map in this inner product is $B^\top/\sqrt m$.
We project the core and expand its coefficient space increment by
\begin{equation}\label{eq:basis_evaluation}
 \widetilde{\cT}_i=\cT_i\times_2(B^\top/\sqrt m),
 \qquad
 \Delta_i=\widetilde\Delta_i\times_2(\sqrt m\,B).
\end{equation}
The local update applies the activation to the expanded increment before
adding it to the input core. The t-attention update adds the expanded
increment directly.

\Needspace{5\baselineskip}
\section{Numerical Reference Checks}\label{sec:reference_checks}
\begingroup
\setlength{\intextsep}{6pt plus 1pt minus 1pt}
\setlength{\abovecaptionskip}{6pt}
\setlength{\belowcaptionskip}{6pt}

\subsection{HJB}
The Cole-Hopf transform $w=\exp(-u/(2\nu))$ reduces \cref{eq:hjb}
to a heat equation. We solve it on grids with 32 and 64 points per
coordinate and invert the transform to obtain $u_{32}$ and $u_{64}$.

For each input, we draw 128 points independently and uniformly from the
experiment grid of 16 points per coordinate and evaluate the dataset target,
$u_{32}$, and $u_{64}$ there. In each column of \cref{tab:hjb_references},
we divide the sampled $L^2$ difference by the norm of the second
solution's sampled values. We report the maximum over all 20,480
training, validation, and test inputs at each size.
\begin{table}[H]
\centering
\small
\caption{HJB reference checks on sampled grid points.}\label{tab:hjb_references}
\setlength{\tabcolsep}{3pt}
\renewcommand{\arraystretch}{1.12}
\begin{tabular*}{\linewidth}{@{\extracolsep{\fill}}lrr@{}}
\toprule
$N$ & Dataset target vs.\ $u_{32}$ & $u_{32}$ vs.\ $u_{64}$ \\
\midrule
4 & $1.62\times10^{-6}$ & $1.56\times10^{-6}$ \\
12 & $2.76\times10^{-5}$ & $4.65\times10^{-6}$ \\
20 & $2.44\times10^{-5}$ & $6.99\times10^{-6}$ \\
28 & $2.80\times10^{-5}$ & $8.31\times10^{-6}$ \\
36 & $2.91\times10^{-5}$ & $1.17\times10^{-5}$ \\
\bottomrule
\end{tabular*}
\end{table}

To quantify truncation error at the model's output rank, we round all
2,048 test targets at $N=4,12,20$ to a maximum bond dimension of ten.
The full-grid mean relative Frobenius errors, computed in TT form, are
$0.0050\%$, $0.00088\%$, and $0.000081\%$, respectively.
The saved test targets at $N=28$ and $N=36$ already meet the output-rank
cap, with maximum bond dimensions of ten and nine, respectively.

\subsection{Ising}
We check the free fermion reference states for \cref{eq:ising}
on two validation inputs at each $N=16,32,64,128,256$. We select one
with the smallest even-sector gap and one with the largest conditional
upper bound on the original reference state's infidelity.

To refine the selected references, we tighten the Gaussian compression
and TT-rounding tolerances from
$2\times10^{-7}$ and $10^{-8}$ to $2\times10^{-9}$ and $10^{-10}$,
respectively, using a bond dimension cap of 64.
\begin{table}[H]
\centering
\small
\caption{Ising reference sensitivity to tighter compression budgets.}\label{tab:ising_references}
\setlength{\tabcolsep}{3pt}
\renewcommand{\arraystretch}{1.12}
\begin{tabular*}{\linewidth}{@{\extracolsep{\fill}}lrr@{}}
\toprule
$N$ & Maximum retained TT ranks & Maximum relative change (\%) \\
\midrule
16 & $10\to12;\ 8\to10$ & 0.0082 \\
32 & $13\to18;\ 12\to16$ & 0.0088 \\
64 & $17\to24;\ 16\to23$ & 0.0093 \\
128 & $23\to30;\ 24\to32$ & 0.0102 \\
256 & $25\to33;\ 46\to58$ & 0.0122 \\
\bottomrule
\end{tabular*}
\par\vspace{3pt}
\begin{minipage}{\linewidth}\footnotesize
Arrows give original-to-refined ranks for the minimum gap input, then the largest bound input.
\end{minipage}
\end{table}

We normalize and phase-align the states in each comparison.
\Cref{tab:ising_references} reports the larger relative $L^2$ change
over the two inputs at each size.

\subsection{Glauber}
For each size, we check the time-evolving block decimation (TEBD)
references on 16 training inputs at evenly spaced indices. On these
inputs, we double the number of time steps at the original bond cap
and increase the bond cap by eight at the original step count.
We compute the full state relative $L^2$ differences in FP64 before
casting the references to FP32. Each denominator is the norm of the
refined reference. All maxima in \cref{tab:glauber_references} are below
$10^{-5}$.
\begin{table}[H]
\centering
\small
\caption{Glauber references and maximum relative changes under TEBD refinement.}\label{tab:glauber_references}
\setlength{\tabcolsep}{3pt}
\renewcommand{\arraystretch}{1.12}
\begin{tabular*}{\linewidth}{@{\extracolsep{\fill}}rrrrr@{}}
\toprule
$N$ & Time steps & Bond cap & Step refinement & Rank refinement \\
\midrule
20 & 32 & 16 & $5.55\times10^{-6}$ & $7.13\times10^{-8}$ \\
40 & 32 & 16 & $9.58\times10^{-6}$ & $1.09\times10^{-7}$ \\
60 & 48 & 16 & $5.55\times10^{-6}$ & $1.27\times10^{-7}$ \\
80 & 48 & 16 & $7.07\times10^{-6}$ & $1.19\times10^{-7}$ \\
100 & 48 & 16 & $8.45\times10^{-6}$ & $7.11\times10^{-8}$ \\
\bottomrule
\end{tabular*}
\end{table}

To compute the nonnegative fraction $f_{\geq 0}$ in
\cref{tab:glauber}, we sample 512 configurations uniformly with
replacement per test input and use the same configurations for both methods.
\par
\endgroup

\Needspace{5\baselineskip}
\section{Ising Training Augmentation}\label{sec:ising_sampling}
For each additional training input at $N=16,64,128,256$, we draw $J_i$
and $\eta_i$ from their base distributions, then draw a target
$d\sim\mathcal N(-0.1,0.1^2)$. To obtain $d_N=d$ in
\cref{eq:ising_distance} with $h_i=\bar h\eta_i$, we set
\begin{equation}\label{eq:ising_sample_scale}
 \bar h=\exp\!\left(
 d-\frac1N\sum_{i=1}^{N}\log\eta_i+
       \frac1{N-1}\sum_{i=1}^{N-1}\log J_i\right).
\end{equation}
If $\bar h\notin[0.5,1.5]$, we redraw $d$ while keeping $J_i$ and $\eta_i$ fixed.
At $N=32$, we sample around $d_N=0$ and multiply each field by
$\exp(-0.1)$ to shift $d_N$ by $-0.1$, with the final $\bar h$ values
in $[0.5,1.5]$.

\section{Canonicalization on a Fixed Tree}\label{sec:treecanon}
We extend \cref{alg:tt_canonicalization} to a labeled tree $G=(V,E)$
with root $\rho$ and fixed orders of the core modes and open indices.
The unfolding $\operatorname{unfold}_{v,e}$ puts mode $k(v,e)$ from
\cref{eq:gauge} in the rows and the other modes in the columns.
Its inverse $\operatorname{fold}_{v,e}$ restores the mode order with
the updated bond dimension.

In \cref{alg:tree_canonicalization}, $S$ contains the remaining vertices
and $v_0$ is the non-root leaf with the smallest label. Labels break
distance ties in $G[S]$. Removing $v$ leaves an open index $\beta_v$ in
its neighbor's former bond slot. We use thin QR and the full row-space
basis in \cref{eq:fixsvd_gram_schmidt}. \Cref{fig:tree_canonicalization}
shows the first removal.

\begin{figure}[H]
\centering
\setlength{\abovecaptionskip}{6pt}
\setlength{\belowcaptionskip}{6pt}
\begingroup
  \colorlet{TreeInk}{black!92}
  \colorlet{TreeBond}{black!68}
  \colorlet{TreeOpen}{teal!70!black}
  \colorlet{TreeBackground}{white}
  \definecolor{TreeQR}{HTML}{365D86}
  \definecolor{TreeCenter}{HTML}{8F602F}
\colorlet{TreeCenterFill}{TreeBackground!76!TreeCenter}
\begin{tikzpicture}[
  every node/.style={font=\small,text=TreeInk},
  vertex/.style={draw=TreeBond,fill=TreeBackground,rounded corners=1.3pt,
    line width=.5pt,minimum width=5.6mm,minimum height=4.2mm,inner sep=1pt},
  root/.style={vertex,minimum width=10.2mm},
  center/.style={draw=TreeCenter,fill=TreeCenterFill,line width=.85pt},
  bond/.style={draw=TreeBond,line width=.6pt},
  qr/.style={draw=TreeQR,line width=.72pt,-{Latex[length=3.8pt,width=3pt]}},
  open/.style={draw=TreeOpen,line width=.72pt},
  index/.style={text=TreeOpen,inner sep=2pt},
  transition/.style={draw=TreeBond,line width=.55pt,-{Latex[length=4pt,width=3.2pt]}},
  stage/.style={font=\footnotesize,inner sep=1pt}
]
\node[root] (a4) at (0,0) {$\rho=4$};
\node[vertex,center] (a1) at (-1.05,-.46) {$1$};
\node[vertex] (a2) at (0,.84) {$2$};
\node[vertex] (a3) at (1.05,-.46) {$3$};
\draw[qr] (a2) -- (a4);
\draw[qr] (a3) -- (a4);
\draw[qr] (a4) -- (a1);
\node[stage] at (0,-.91) {QR toward leaf $1$};

\begin{scope}[xshift=4.1cm]
  \node[root,center] (b4) at (0,0) {$\rho=4$};
  \node[index] (b1) at (-1.05,-.46) {$\beta_1$};
  \node[vertex] (b2) at (0,.84) {$2$};
  \node[vertex] (b3) at (1.05,-.46) {$3$};
  \draw[open] (b4) -- (b1);
  \draw[bond] (b4) -- (b2);
  \draw[bond] (b4) -- (b3);
  \node[stage] at (0,-.91) {Remove leaf $1$};
\end{scope}

\begin{scope}[xshift=8.2cm]
  \node[root] (c4) at (0,0) {$\rho=4$};
  \node[index] (c1) at (-1.05,-.46) {$\beta_1$};
  \node[vertex,center] (c2) at (0,.84) {$2$};
  \node[vertex] (c3) at (1.05,-.46) {$3$};
  \draw[open] (c4) -- (c1);
  \draw[qr] (c4) -- (c2);
  \draw[bond] (c4) -- (c3);
  \node[stage] at (0,-.91) {QR toward leaf $2$};
\end{scope}

\draw[transition] (1.65,.26) -- (2.45,.26);
\draw[transition] (5.75,.26) -- (6.55,.26);
\end{tikzpicture}
\endgroup
\caption{Initial QR sweep and first leaf removal. Shading marks the orthogonality center.}
\label{fig:tree_canonicalization}
\end{figure}
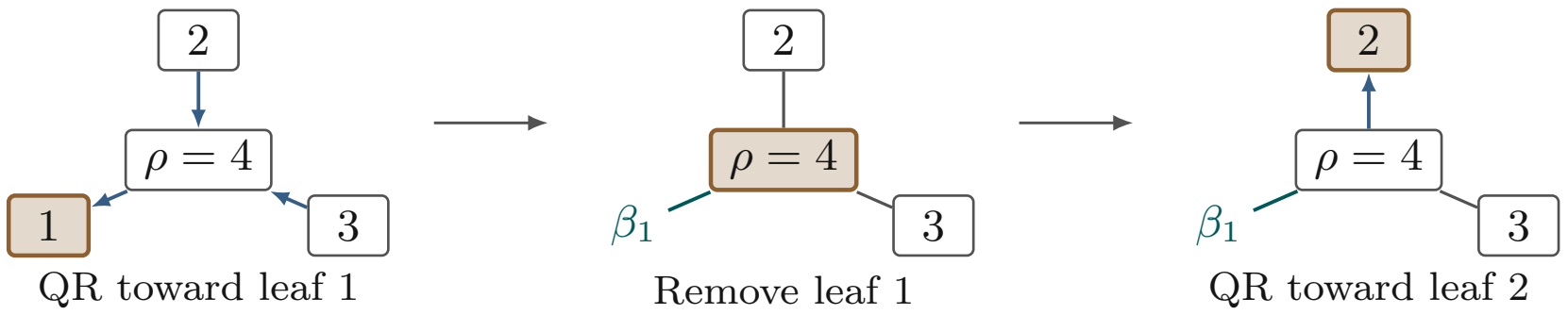

A one-vertex tree returns its sole core. A zero contraction returns
$\mathbf{0}_{\mathrm{tree}}$, with unit bond dimensions, a zero root
core, and $e_1$ at each other vertex after grouping its open modes.

\begin{algorithm}[H]
\caption{Tree canonicalization $\Can_{\mathrm{tree}}$}
\label{alg:tree_canonicalization}
\begin{algorithmic}[1]
\Require Real cores $\tn=(\cT_v)_{v\in V}$ with a nonzero contraction and $|V|>1$.
\Procedure{QRMove}{$p,q$}
  \State $e\gets(p,q)$,\quad
    $(Q,R)\gets\operatorname{qr}\bigl(\operatorname{unfold}_{p,e}(\cT_p)^\top\bigr)$
  \State $\cT_p\gets\operatorname{fold}_{p,e}(Q^\top)$,\quad
    $\cT_q\gets\cT_q\times_{k(q,e)}R$
\EndProcedure
\State $S\gets V$,\quad $v\gets v_0$
\For{$p\in V\setminus\{v_0\}$ in decreasing distance from $v_0$}
  \State $q\gets$ the next vertex on the path from $p$ to $v_0$
  \State \Call{QRMove}{$p,q$}
\EndFor
\While{$|S|>1$}
  \State $u\gets$ the neighbor of $v$ in $G[S]$,\quad $e\gets(v,u)$
  \State $T_v\gets\operatorname{unfold}_{v,e}(\cT_v)$,\quad
    $W_v\gets\operatorname{FixSVD}(T_v)$
  \State $\cT_v^{\rm can}\gets\operatorname{fold}_{v,e}(W_v^\top)$,\quad
    $\cT_u\gets\cT_u\times_{k(u,e)}(T_vW_v)^\top$
  \State $S\gets S\setminus\{v\}$, exposing $\beta_v$ in mode $k(u,e)$
  \If{$|S|>1$}
    \State $v\gets$ the non-root leaf of $G[S]$ nearest to $u$
    \For{$(p,q)$ along the path from $u$ to $v$}
      \State \Call{QRMove}{$p,q$}
    \EndFor
  \EndIf
\EndWhile
\State $\cT_\rho^{\rm can}\gets\cT_\rho$
\State \Return $(\cT_v^{\rm can})_{v\in V}$
\end{algorithmic}
\end{algorithm}

On the path $1-\cdots-N$ rooted at $1$, the sweep toward $N$ and
the leaf removal order $N,\ldots,2$ recover \cref{alg:tt_canonicalization}.

\begin{theorem}[Tree canonicalization]\label{thm:tree_canonical}
\upshape
Fix the tree, root, and index orders in \cref{alg:tree_canonicalization}.
If two core tuples $\tn$ and $\tn'$ on the
tree represent the same tensor $\cT$, then
\begin{equation}\label{eq:tree_canonicalization_identity}
 \Can_{\mathrm{tree}}(\tn)=\Can_{\mathrm{tree}}(\tn'),\qquad
 \jointcontract_E\bigl(\Can_{\mathrm{tree}}(\tn)\bigr)=\cT .
\end{equation}
Equality of canonicalizations holds core by core, including bond dimensions.
\end{theorem}

\begin{proof}
The zero and one-vertex cases follow from the output conventions.
At a vertex distinct from the center, the unfolding toward the center
uses the first edge of the path between them. A QR move from $p$ to $q$
preserves the contraction and sets the unfolding of $\cT_p$ along
$(p,q)$ to $Q^\top$, whose rows are orthonormal. The initial sweep
establishes this property toward $v_0$. Removing a leaf preserves it
toward the absorbing neighbor, and the next QR moves carry the center
to the next leaf.

At the current leaf $v$, let $P_v$ be the contraction of the other
cores in $G[S]$, with the bond to $v$ as its column index. Each
branch map composes the transpose of its core unfolding with the
tensor product of its child branch maps and the identity on its
open modes. The transpose has orthonormal columns by the QR argument,
and the child maps do by induction. Tensor
products and compositions preserve column orthonormality, so
$P_v^\top P_v=I$.

\Needspace{3\baselineskip}
For two inputs representing the same tensor, the initial sweeps leave
the same contraction and no saved cores. Inductively, assume the
remaining tensor and saved cores agree. Unfold the remaining tensor
as $M_v$ with the open modes at $v$ in its columns. In each run,
\begin{equation}\label{eq:tree_remaining}
 M_v=P_vT_v,\qquad T_v^\top T_v=M_v^\top M_v.
\end{equation}
By \cref{eq:fixsvd_gram_schmidt}, the common Gram matrix determines
the same $W_v$ and saved leaf core. The update at $u$ gives the next
remaining tensor by reshaping $M_v W_v$ with $\beta_v$ in the former
bond slot. Since $W_v$ spans the row space of $M_v$,
\begin{equation}\label{eq:tree_row_space}
 M_v=(M_vW_v)W_v^\top .
\end{equation}
Thus the next tensor is nonzero, and reattaching the leaf is exact.
Both runs choose the same next leaf, and QR moves preserve the common
remaining tensor. Induction gives the same root core, and reattaching
the saved cores recovers $\cT$.
\end{proof}
\section{TT-CNN on DSADS}\label{sec:dsads_details}
We divide each 125-point input window into 25 consecutive five-point
blocks. Four convolutional layers process the block sequence with
$3$, $6$, $11$, and $22$ output channels, respectively. Pooling and
a linear readout produce the 19 class scores.
For one TT-CNN convolution in \cref{sec:dsads}, write $X_{k,c,q,t}$
for an input entry and $Y_{k,c',q,u}$ for an output entry before batch
normalization. The index $k$ labels the time block, $c$ and $c'$ label the
input and output feature channels, and $t,u\in\{1,\ldots,5\}$ index time
within a block. The index $q$ identifies one of the 45 measurements by its
body location, sensor modality, and spatial direction. With
$C_{\mathrm{in}}$ input feature channels, the convolution in
\cref{fig:dsads_ttcnn_convolution} is
\begin{equation}\label{eq:dsads_core_convolution}
 Y_{k,c',q,u}
 =\sum_{\delta=-1}^{1}\sum_{c=1}^{C_{\mathrm{in}}}\sum_{t=1}^{5}
 W_{c',c,\delta,u,t}\,X_{k+\delta,c,q,t}.
\end{equation}
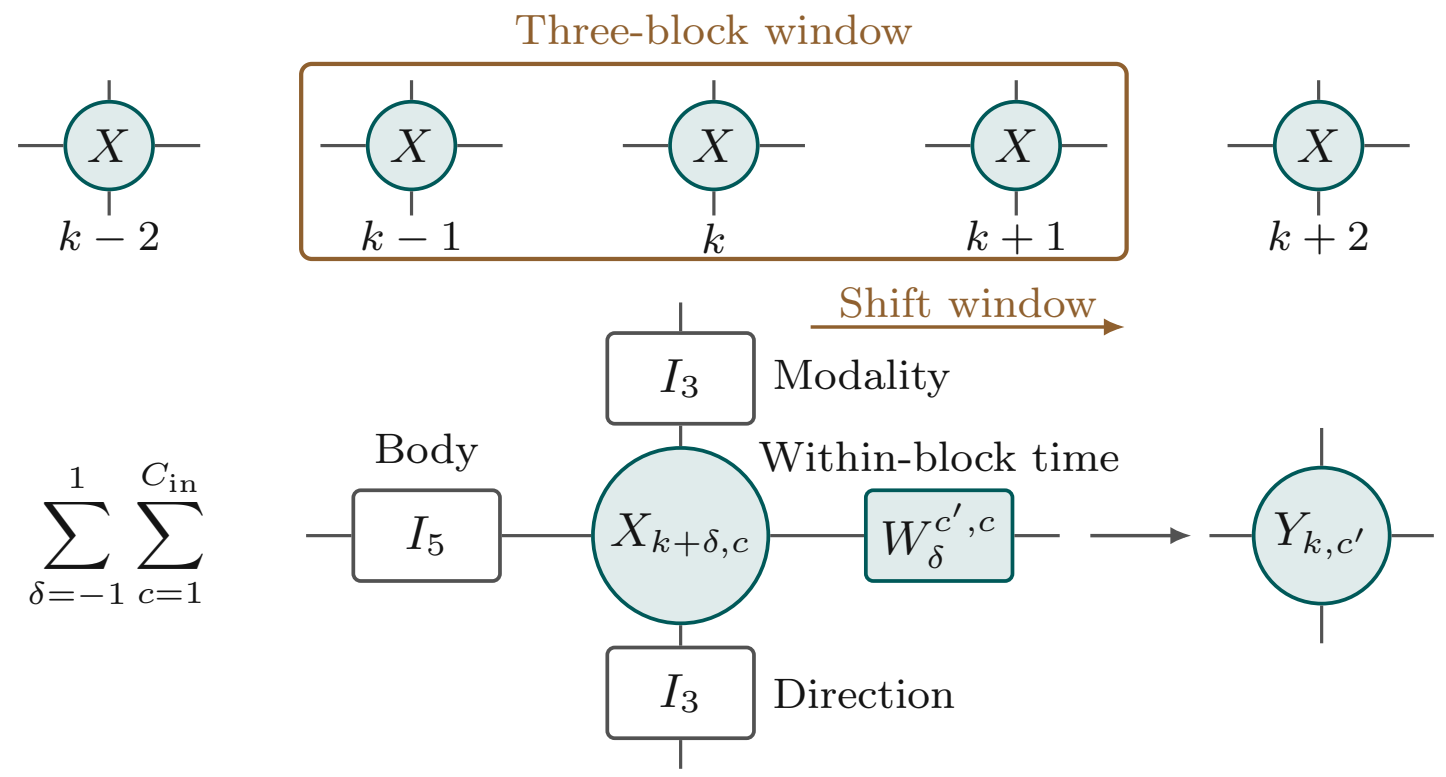
\begin{figure}[!htbp]
\centering
\setlength{\abovecaptionskip}{6pt}
\setlength{\belowcaptionskip}{6pt}
\begingroup
  \colorlet{TTConvInk}{black!92}
  \colorlet{TTConvBond}{black!68}
  \colorlet{TTConvMeasurement}{teal!70!black}
  \colorlet{TTConvPaper}{white}
  \definecolor{TTConvWindow}{HTML}{8F602F}
\colorlet{TTConvCoreFill}{TTConvPaper!88!TTConvMeasurement}
\begin{tikzpicture}[
  every node/.style={font=\small,text=TTConvInk},
  core/.style={circle,draw=TTConvMeasurement,fill=TTConvCoreFill,
    line width=.7pt,minimum size=5.8mm,inner sep=2pt},
  kernel/.style={draw=TTConvMeasurement,fill=TTConvCoreFill,rounded corners=1.4pt,
    line width=.7pt,minimum width=9.7mm,minimum height=6mm,inner sep=2pt},
  identity/.style={kernel,draw=TTConvBond,fill=TTConvPaper},
  bond/.style={draw=TTConvBond,line width=.6pt},
  annotation/.style={font=\footnotesize,inner sep=1.5pt},
  window/.style={draw=TTConvWindow,rounded corners=2pt,line width=.8pt},
  advance/.style={draw=TTConvBond,line width=.65pt,-{Latex[length=4pt,width=3.3pt]}}
]
\draw[window] (2.27,1.03) rectangle (7.73,2.32);
\node[annotation,text=TTConvWindow] at (5,2.55) {Three-block window};
\foreach \site/\xpos/\blockindex in {
  lefttwo/1/{k-2},leftone/3/{k-1},block/5/k,rightone/7/{k+1},righttwo/9/{k+2}}{
  \node[core] (\site) at (\xpos,1.78) {$X$};
  \draw[bond] (\site.west) -- ++(-.30,0);
  \draw[bond] (\site.east) -- ++(.30,0);
  \draw[bond] (\site.north) -- ++(0,.13);
  \draw[bond] (\site.south) -- ++(0,-.16);
  \node[annotation] at (\xpos,1.17) {$\blockindex$};
}
\draw[advance,draw=TTConvWindow] (5.64,.58) --
  node[annotation,above,text=TTConvWindow] {Shift window} (7.72,.58);

\node at (1.05,-.80) {$\displaystyle\sum_{\delta=-1}^{1}\sum_{c=1}^{C_{\mathrm{in}}}$};
\node[core] (input) at (4.78,-.80) {$X_{k+\delta,c}$};
\node[identity] (body) at (3.10,-.80) {$I_5$};
\node[identity] (modality) at (4.78,.24) {$I_3$};
\node[identity] (direction) at (4.78,-1.84) {$I_3$};
\node[kernel] (time) at (6.49,-.80) {$W_\delta^{c',c}$};
\draw[bond] (input) -- (body);
\draw[bond] (input) -- (modality);
\draw[bond] (input) -- (direction);
\draw[bond] (input) -- (time);
\draw[bond] (body.west) -- ++(-.30,0);
\draw[bond] (modality.north) -- ++(0,.19);
\draw[bond] (direction.south) -- ++(0,-.19);
\draw[bond] (time.east) -- ++(.30,0);
\node[annotation,above=1.5pt] at (body.north) {Body};
\node[annotation,right=1.5pt] at (modality.east) {Modality};
\node[annotation,right=1.5pt] at (direction.east) {Direction};
\node[annotation,above=1.5pt] at (time.north) {Within-block time};

\draw[advance] (7.49,-.80) -- (8.16,-.80);
\node[core] (output) at (9.02,-.80) {$Y_{k,c'}$};
\draw[bond] (output.west) -- ++(-.28,0);
\draw[bond] (output.east) -- ++(.28,0);
\draw[bond] (output.north) -- ++(0,.25);
\draw[bond] (output.south) -- ++(0,-.25);
\end{tikzpicture}
\endgroup
\caption{TT-CNN convolution at block $k$.}
\label{fig:dsads_ttcnn_convolution}
\end{figure}

The convolution applies identity matrices to the three measurement
modes represented by $q$ and the matrix
$W_\delta^{c',c}=(W_{c',c,\delta,u,t})_{u,t}$ to the within block time
mode (\cref{fig:dsads_ttcnn_convolution}).
We pad the sequence with one zero block at each end. Max pooling along
$k$ uses length-two windows and stride two. When the block sequence has
odd length, the last pooling window contains only the final block.

The 1D CNN uses the 45 measurements as input channels. Its four
convolutional layers use length-three kernels along the original time
axis, with $32,64,128,256$ output channels.

We use HAROOD's preprocessed windows with two held-out subjects per
fold. Within each source domain, we use a stratified 80/20 training and
validation split. Each fold contains 5,472 training windows, 1,368
validation windows, and 2,280 test windows. We train both CNNs for 150
epochs using cross-entropy loss, Adam with a learning rate of
$10^{-3}$, and batch size 128. If checkpoints for either CNN tie on
accuracy over the combined source-domain validation set, we choose
the one with the lowest cross-entropy on that set.

\begin{table}[H]
\centering
\small
\caption{DSADS test accuracy by held-out subject pair.}\label{tab:dsads_folds}
\begin{tabular*}{\linewidth}{@{\extracolsep{\fill}}lrr@{}}
\toprule
Held-out subjects & 1D CNN (\%) & TT-CNN (\%) \\
\midrule
p8, p2 & 85.00 & 89.56 \\
p1, p3 & 86.75 & 87.19 \\
p7, p6 & 89.04 & 95.26 \\
p4, p5 & 74.47 & 93.38 \\
\bottomrule
\end{tabular*}
\end{table}

\Cref{tab:dsads_folds} uses the original DSADS subject identifiers.
The URM accuracy in \cref{tab:dsads} comes from Table~10 of the
extended HAROOD paper. We estimate its parameter count from the public
implementation's input projection, transformer encoder blocks, and classifier.
\clearpage
\endgroup
\end{document}